\documentclass{article} % For LaTeX2e
\usepackage{gmas_iclr2022_conference,times}
\usepackage{amsmath,amsfonts,bm}

\def\eqref#1{equation~\ref{#1}}
\def\1{\bm{1}}

\def\va{{\bm{a}}}

\def\vs{{\bm{s}}}

\def\vv{{\bm{v}}}

\def\vz{{\bm{z}}}

\def\eva{{a}}

\def\evc{{c}}
\def\evd{{d}}

\def\evr{{r}}
\def\evs{{s}}

\def\evu{{u}}
\def\evv{{v}}
\def\evw{{w}}

\DeclareMathAlphabet{\mathsfit}{\encodingdefault}{\sfdefault}{m}{sl}
\SetMathAlphabet{\mathsfit}{bold}{\encodingdefault}{\sfdefault}{bx}{n}

\def\sP{{\mathbb{P}}}

\def\sV{{\mathbb{V}}}

\DeclareMathOperator*{\argmax}{arg\,max}

\usepackage{hyperref}
\usepackage{url}
\usepackage{graphicx, wrapfig}
\usepackage{subcaption}
\usepackage{comment}
\usepackage{optidef}
\usepackage{amsmath}
\usepackage{ulem}
\usepackage{multirow, multicol}

\title{Multi-Agent Neural Rewriter for Vehicle Routing with Limited Disclosure of Costs}

\author{Nathalie Paul \\
Fraunhofer Center for Machine Learning \\
Fraunhofer IAIS\\
\texttt{nathalie.paul@iais.fraunhofer.de}
\And
Tim Wirtz \\
Fraunhofer Center for Machine Learning\\
Fraunhofer IAIS\\
\texttt{tim.wirtz@iais.fraunhofer.de}
\AND
Stefan Wrobel \\
University of Bonn \\
Fraunhofer Center for Machine Learning \\
Fraunhofer IAIS \\
\texttt{stefan.wrobel@cs.uni-bonn.de}
\And
Alexander Kister \\
eScience Division (S.3) \\
Federal Institute for Materials Re-\\search and Testing\\
\texttt{alexander.kister@bam.de}}

\definecolor{nice_green}{rgb}{0.0, 0.62, 0.42}

\iclrfinalcopy

\begin{document}

\maketitle

\begin{abstract}
We interpret solving the multi-vehicle routing problem as a team Markov game with partially observable costs. For a given set of customers to serve, the playing agents (vehicles) have the common goal to determine the team-optimal agent routes with minimal total cost. Each agent thereby observes only its own cost.
Our multi-agent reinforcement learning approach, the so-called multi-agent Neural Rewriter, builds on the single-agent Neural Rewriter to solve the problem by iteratively rewriting solutions.
Parallel agent action execution and partial observability require new rewriting rules for the game. We propose the introduction of a so-called pool in the system which serves as a collection point for unvisited nodes. It enables agents to act simultaneously and exchange nodes in a conflict-free manner.
We realize limited disclosure of agent-specific costs by only sharing them during learning. During inference, each agents acts decentrally, solely based on its own cost.
First empirical results on small problem sizes demonstrate that we reach a performance close to the employed OR-Tools benchmark which operates in the perfect cost information setting. 

\end{abstract}
\section{Introduction}
\begin{wrapfigure}{r}{0.35\textwidth}
    \includegraphics[width=0.35\textwidth]{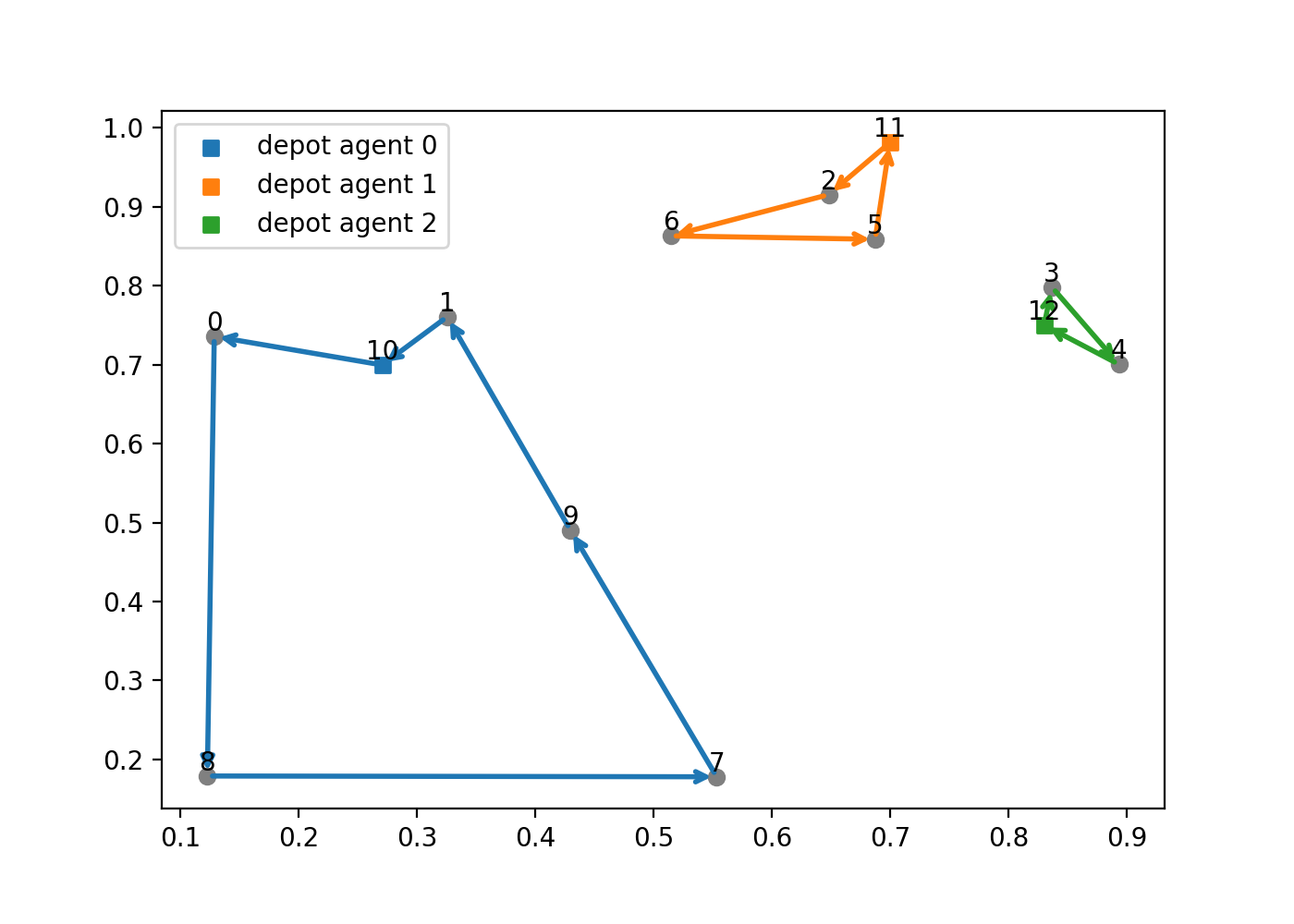}
    \caption{An exemplary solution of a vehicle routing problem with $10$ customers and $3$ vehicles: Each customer is represented by a node. Each vehicle has its own depot (starting node). A vehicle's route is given by a sequence of visited nodes. Each edge induces a vehicle-specific cost.}
    \label{intro_vrp}
\end{wrapfigure}

While the logistics market is typically determined by strongly competing players, there also exist initiatives to encourage collaboration for, e.g., more cost-efficient and sustainable resource usage\footnote{https://eshipco.com/en/}\footnote{https://www.transporeon.com/en/reports/horizontal-collaboration}.
A simplified model of the logistics market is given by a set of customers and a set of vehicles belonging to different logistics companies with vehicle-specific costs. 
The setup of cooperating companies, which seek to jointly optimize the overall costs while serving all customers, is formalized by the (multi-)vehicle routing problem \citep{laporte1992vehicle} (cf.\ Figure \ref{intro_vrp}). It involves assigning customers to the companies as well as determining the corresponding vehicle routes.
A natural additional constraint is that companies hereby want to keep certain types of their costs private. E.g., company-specific driver costs, fuel consumptions and also fixed costs for marketing or customer service, reflect business-internal information. Sharing such information would lead to a decisive competitive disadvantage.
We interpret solving the multi-vehicle routing problem as a team Markov game \citep{wang2002reinforcement} with partially observable 
costs: The parallel acting vehicles play a cooperative game to build the (ideally) team-optimal vehicle routes, whereby each vehicle can observe only its own local cost information.
While other approaches for collaborative vehicle routing have studied the scenario of self-interested vehicles (companies) which can form coalitions involving only a subset of vehicles \citep{Mak2021}, we focus on the setup of selfless, team-oriented vehicles which all participate in the collaboration.
As reinforcement learning (RL) has lately become very attractive for learning heuristics for NP-hard combinatorial optimization problems like vehicle routing problems \citep{nazari2018reinforcement,kool2019attention,chen2019learning}, we build on an existing RL framework to implement our game. The so-called Neural Rewriter \citep{chen2019learning} considers the setup of a single vehicle and learns how to iteratively improve a given solution within a rewriting episode of fixed length. In each step of the episode the current routing solution, given by a sequence of visited nodes, gets slightly changed by swapping two nodes. Our proposed multi-agent Neural Rewriter (MANR) extends the Neural Rewriter to a multi-agent system with multiple vehicles (agents) with individual costs. We assume the individual cost matrices to originate from the same underlying distribution, i.e., we consider a set of homogeneous agents.
To ensure the required parallel action execution in the team Markov game, we have to avoid conflicting rewriting actions of agents. The assumption of partially observable costs poses additional requirements on the rewriting game structure: Given the incomplete knowledge about its team members, an agent must never modify the routes of other agents but can solely alter its own local route. Yet at the same time, agents must be able to exchange nodes, i.e., to change the customer-company-assignment. This raises the question: How should agents know whether excluding a customer node or integrating a new customer node is beneficial for the team if they only observe their own costs? There is a trade-off between non-disclosure and optimality; some sort of information exchange between the team members is inevitable for finding a global optimum.
We solve the described challenges on different levels of the MANR.
On the game setup level, the key is the introduction of a pool set to coordinate agent actions and prevent conflicts. It serves as a collection point for agents to drop customer nodes and take customer nodes from. 
Providing the local agents with some global knowledge for deciding about exchanging customer nodes is ensured by the employed learning approach. 
We ultimately seek to limit the necessary exchange of cost information between the players as much as possible. As a first step, we realize limited disclosure of vehicle-specific costs only during inference and consider shared costs during training. This is a valid assumption from an application perspective as training can be performed based on fictitious company cost information sampled from a realistic cost distribution. Our actor-critic setup follows the idea of centralized-learning-decentralized-execution \citep{zhang2021multi}: During training, all vehicle cost information is shared to learn a centralized critic for estimating the expected team benefit of rewriting actions. The critic passes this global knowledge on to an agent policy, which itself observes only local information. An agent thus learns to locally behave as a team player, assuming a representative team. During inference, each agent uses only his local cost (and not the cost of other agents) to determine his actions.

We give an overview of the related work in Section \ref{rel_work}. 
In Section \ref{MANR} we define the considered collaborative vehicle routing problem and present and discuss the adapted Neural Rewriter, the so-called multi-agent Neural Rewriter as a solution approach. Section \ref{sec_emp_eval} empirically evaluates the MANR on simulated data for different setups varying the size of the routing problem as well as the number of vehicles and compares its performance to OR-Tools. Section \ref{conclusion} summarizes the results and outlines plans for future work.

\section{Related Work}
\label{rel_work}
\textbf{RL for vehicle routing}\hspace{0.2cm} RL has been successfully used to tackle NP-hard vehicle routing problems. Most of the work focuses on setups involving a single vehicle. 
E.g., \citet{chen2019learning} consider a capacitated vehicle routing problem where a single vehicle has to visit a set of customers in (possibly) multiple tours starting and ending at its depot without exceeding the vehicle's capacity within one tour. Their approach follows the idea of local search, a well-established heuristic solution approach in the operations research community. Based on an initial solution, they iteratively create locally rewritten solutions via local modifications. They define a local modification as swapping two nodes in the current routing solution sequence and let a RL algorithm learn how to create good neighbouring solutions therewith. Other RL-based approaches for vehicle routing typically build a solution sequentially by visiting one customer more in each step of the episode. E.g., both \citet{kool2019attention,nazari2018reinforcement} consider multiple variants of vehicle routing problems which however also concern only a single vehicle. \citet{nazari2018reinforcement} describe the idea of multiple vehicles in a collaborative or competetive setting as an interesting direction and already mention the need to solve emerging conflicts between multiple vehicles. We decided to build on the rewriting approach of \citet{chen2019learning} as it allows to handle conflicts more naturally. In an iteratively built solution agents either need to perform their actions sequentially to avoid conflicts, which would violate our assumption of a Markov game with simultaneously acting agents, or they need a (stochastically influenced) a posterio conflict solver which would be challenging for agents to learn about. The rewriting approach allows to introduce a pool mechanism for ensuring conflict-free parallel agent actions (details are given below). Moreover, the rewriting setup avoids a challenging sparse reward environment: At each step the cost of the current solution indicates the quality of this solution.\newline\newline
\textbf{Limited disclosure in multi-agent RL}\hspace{0.2cm} In our setup, agents seek to keep their local costs private, whereby these costs influence the reward and also local state representations.
Limiting the disclosure of such agent-specific information is an active research topic for both cooperative and competitive multi-agent systems.  The extreme case of independent learning \citep{matignon2012independent} considers agents as independent entities which observe solely their local environment and are not allowed to share information at all. Other approaches allow for some sort of information exchange as it is known to generally improve performance over independent learning for cooperative tasks \citep{tan1993multi}.
In centralized-learning-decentralized-execution schemes \citep{zhang2021multi, lowe2017multi}, one typically assumes perfect global knowledge about the agents' local states, actions and rewards during training. This information is leveraged to train policies which themselves observe solely local information. For execution, only the policies are used in a decentral manner and thus guarantee limited disclosure during inference.
The literature discusses also a stricter interpretation of limited disclosure by additionally requiring it during the training phase. One possible approach is to share solely locally learned model parameters instead of the local raw data in spirit of distributed machine learning \citep{mcmahan2017communication}. E.g., in \citet{zhang2018fully}, the cooperating agents jointly learn a global value function by sharing the agents' individual parameters of their local estimate of the global value function over a time-varying communication network. 

As a first step, we consider the scenario of limited disclosure during inference, realized by a centralized-learning-decentralized-execution approach. To the best of our knowledge, there exists no RL-based approach to solve a multi-vehicle routing problem with simultaneously acting cooperating agents which can observe solely their own individual cost.

\section{Multi-Agent Neural Rewriter}
\label{MANR}
The developed multi-agent Neural Rewriter (MANR) uses building blocks from the original Neural Rewriter \citep{chen2019learning} to solve a multi-vehicle routing problem with vehicle-specific costs and depot locations in spirit of a team Markov game with partially observable costs.
During conception of the multi-agent setup, we considered the following guiding principles:
$\vspace{-5pt}$
\begin{itemize}
    \item Self-determined agents: Whether and how an agent's route is changed can be decided only by the agent itself and not by other agents. This is essential since, due to the partial cost observability, the agents do not have the necessary information for deciding about optimal routes of other agents.
    \item Conflict-free parallel decisions: Any decision of an agent cannot be in conflict with other agent decisions at the same time step. 
    $\vspace{-5pt}$
\end{itemize}

A naive extension of the Neural Rewriter to multiple vehicles by retaining the original action (and state) space locally for single agents does not satisfy these criteria: An agent swapping any two customer nodes in the overall solution can result in changing another agent's route and also gives room for contradicting agent actions. Solving these issues by simply restricting agents to swap only customer nodes in their own routes would prohibit the necessary customer exchange between agents. Hence, we needed to redefine the rules for our team Markov game.
The approach we propose is the introduction of a pool set over which single agents can get rid of customer nodes or integrate new ones. Agents can interact with each other only via the pool, a direct interaction is not possible.
The pool coordinates the node exchange and with this guarantees the self-determination of agents as well as the conflict-free agent decisions which will be described below in more detail. "Agentizing" the Neural Rewriter also required adaptions of the used models. For a detailed listing of modifications the reader is referred to Appendix \ref{app_diffNR}.
In the following, we first introduce the problem statement in Section \ref{problemStatement} and our translation to a team Markov game in Section \ref{RLSetup}. The implementation of the game with the corresponding MANR model components is described in Section \ref{models}.

\subsection{Problem Statement}
\label{problemStatement}
We consider a multi-vehicle routing problem where $\displaystyle n$ vehicles (agents), characterized by individual costs and depots, collaborate on serving a set of customers $\displaystyle \sV$. Each customer node $\displaystyle \vv \in \sV$ has to be visited exactly once in total by any of the agents. Each agent route starts and ends at its own depot.
A collection of agent routes which satisfies these two criteria is called a feasible solution. The goal is to find an optimal solution which is a feasible solution and has minimal team average costs (cf.\ Figure \ref{intro_vrp}). The team average cost is given by the average over all agent route costs.

\subsection{Team Markov Game}
\label{RLSetup}
\begin{figure}[h]
    \centering
        \raisebox{-\height}{\includegraphics[width=0.32\textwidth]{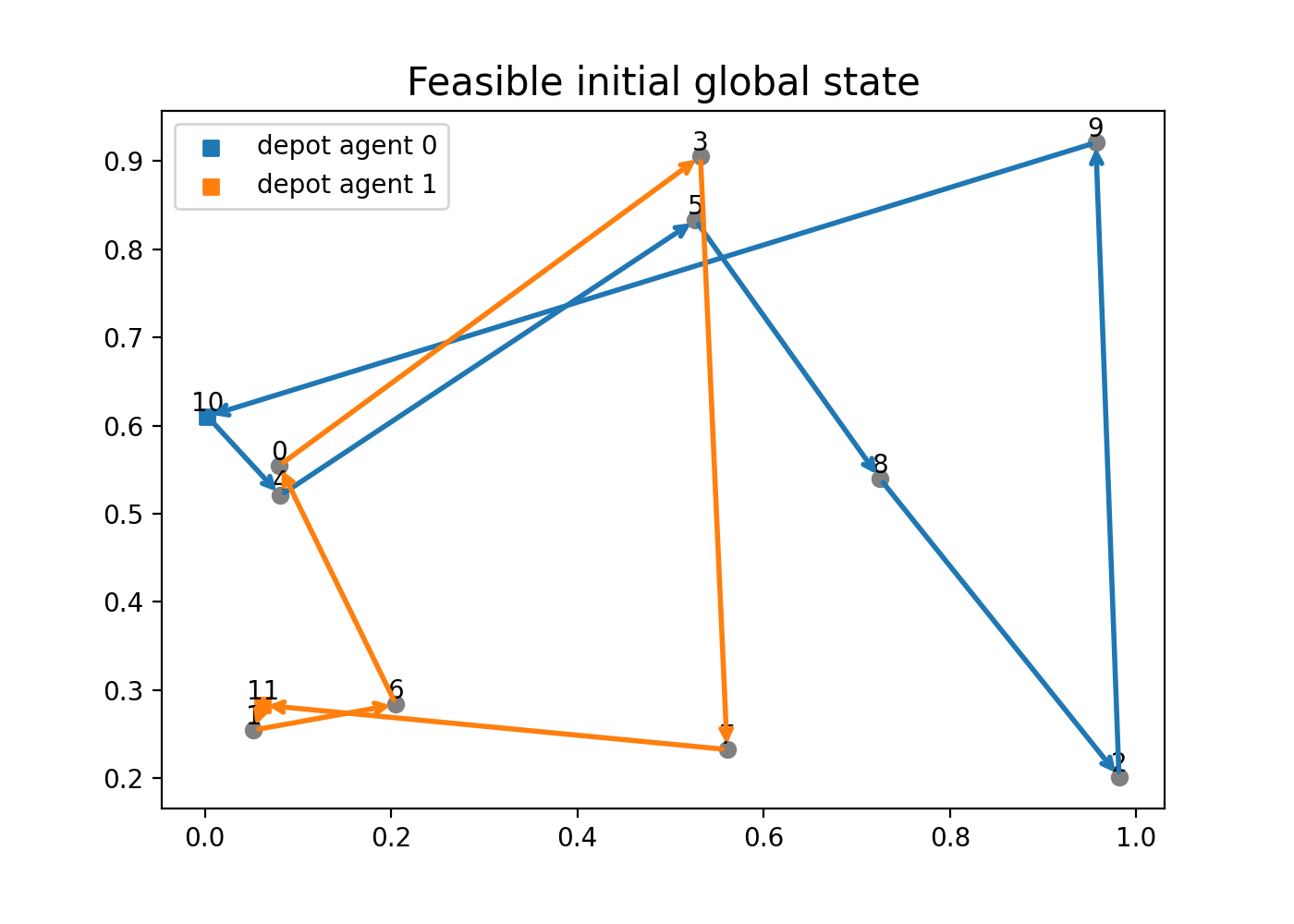}}
        \raisebox{-\height}{\includegraphics[width=0.32\textwidth]{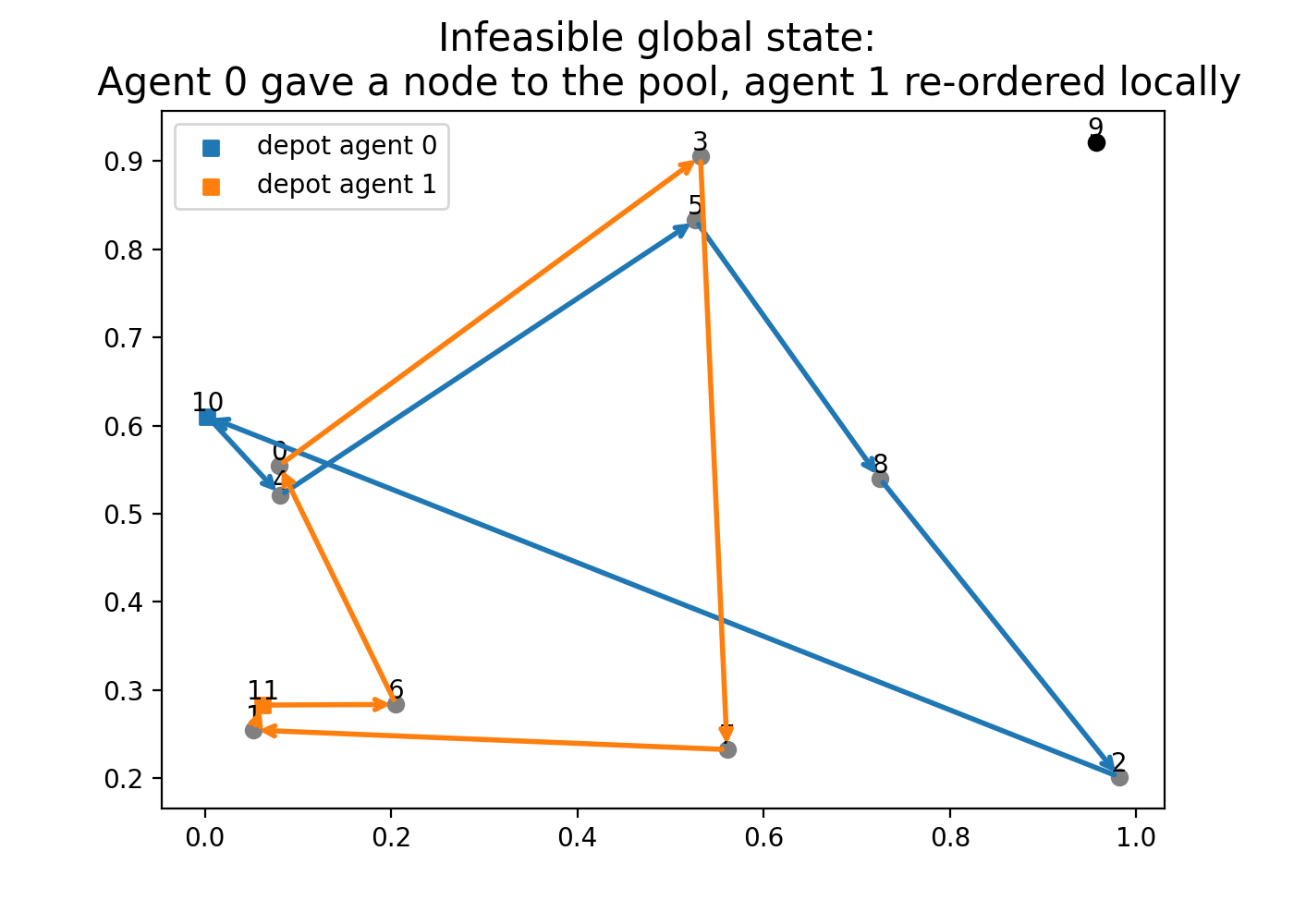}}
        \raisebox{-\height}{\includegraphics[width=0.32\textwidth]{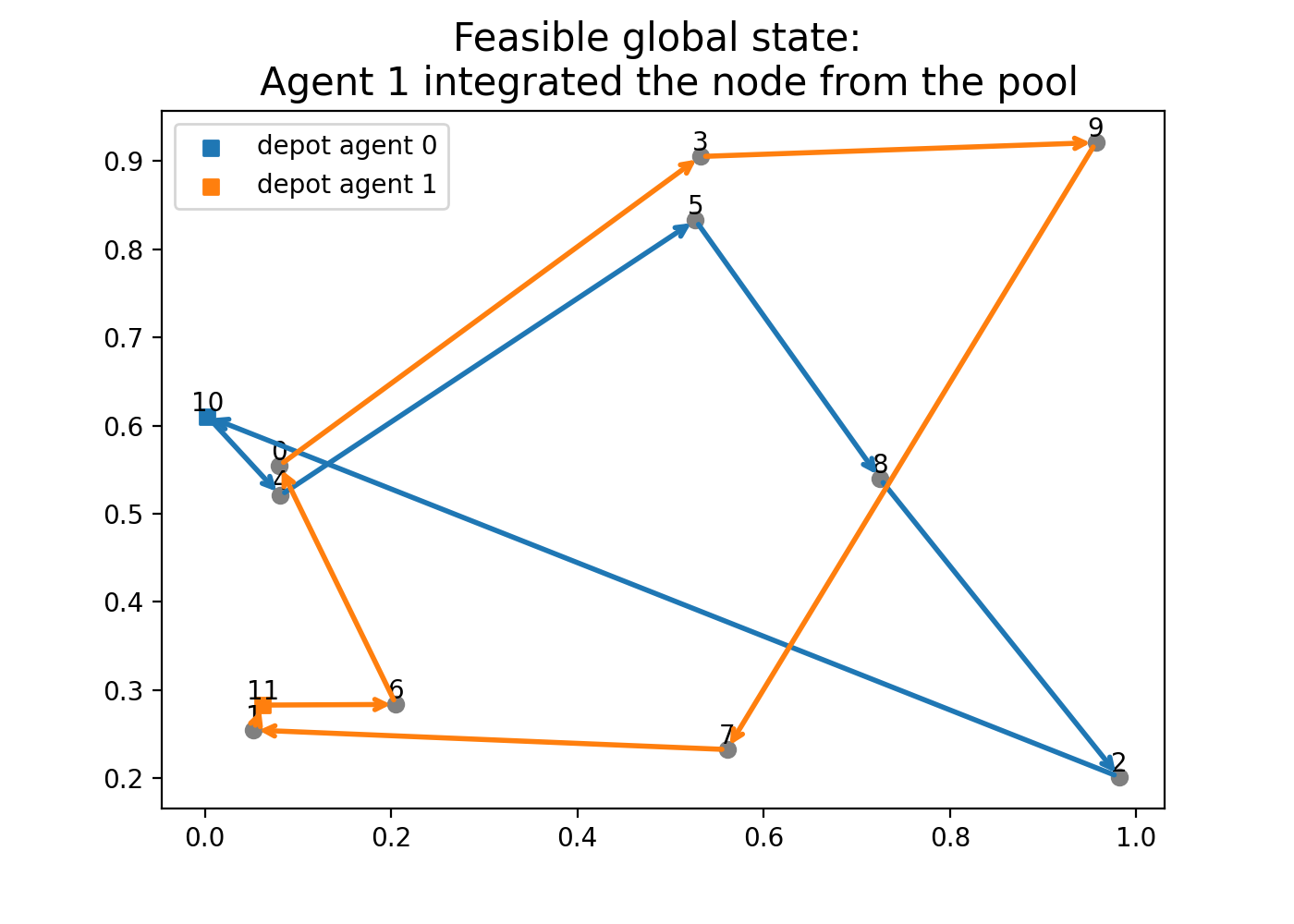}}
    \caption{Exemplary sequence of three global states with corresponding semantic global actions.}
    \label{fig_rewritten_states}
\end{figure}

\citet{chen2019learning} modelled the rewriting procedure of local search for a one-vehicle routing problem as a Markov decision process: States represent routing solutions, actions local modifications of a solution and rewards indicate the cost improvement of a local modification. In our multi-agent system we generally differentiate between two perspectives on the problem: the local perspectives of single agents referring to their own local states and actions and the global perspective which observes all agents' states and actions simultaneously. We consider a single global episode which involves rewriting global states with global actions where all agents execute one local action in parallel. The system obtains global rewards which reflect the overall success of rewriting steps for the whole agent team. To address the challenge of non-conflicting parallel local agent actions, we establish the pool set and model it as an additional component in the system's global state. The pool offers the agents the opportunity to drop customer nodes which they want to exclude from their routes and also to integrate new customer nodes. It also plays an important role for the global team reward as improper usage of the pool leads to a collective penalty.
The precise Markov game setup including the kinds of agent interaction with the pool is described in the following.
\newline\newline
\textbf{States}\hspace{0.2cm} A global state $\displaystyle \vs_t = (\displaystyle \evs^1_t, \displaystyle \evs^2_t, ...,\displaystyle \evs^n_t,\sP_t)$ at time $\displaystyle t$ is defined by the concatenation of all local agent states at time $\displaystyle t$ and the corresponding current state of the pool $\displaystyle \sP_t$. A local state $\displaystyle \evs^i_t$ of agent $\displaystyle i$ at time $\displaystyle t$ is given by the agent's route at time $\displaystyle t$ characterized by the sequence of visited nodes starting and ending at the agent's depot. The pool state $\displaystyle \sP_t$ is a set of nodes which is either empty or contains customer nodes which were dropped there by agents and which are thus unvisited at that time step. We note that not all global states are necessarily feasible solutions to the routing problem but only those with an empty pool. See Figure \ref{fig_rewritten_states} for exemplary global states.

\textbf{Actions}\hspace{0.2cm}
A global action $\displaystyle \va_t = (\displaystyle \eva^1_t, \displaystyle \eva^2_t, ...\displaystyle \eva^n_t)$ at time $\displaystyle t$ is defined by the concatenation of all local agent actions at time $\displaystyle t$. A local action $\displaystyle \eva^i_t$ of agent $\displaystyle i$ at time $\displaystyle t$ involves making two successive decisions which specify the region of the solution to be changed as well as the rule of how to change it. Technically, it means to select a region node $\displaystyle \evw^i_t$ and afterwards a corresponding rule node $\displaystyle \evu^i_t$ with $\displaystyle \eva^i_t = (\displaystyle \evw^i_t, \displaystyle \evu^i_t)$ implying the region node to be moved by being placed after the rule node.
We allow local actions only to rewrite the agent's own local state and update the pool state. Semantically, an agent can re-arrange one node within its local state, integrate a new node by taking it from the pool, exclude a node by giving it to the pool or also keep its local state unchanged (cf.\ Figure \ref{fig_rewritten_states} for examples).
To encourage the frequent generation of feasible solutions, we restrict the set of allowed local actions dependent on the pool state: Each time the pool is filled and hence the global state is an infeasible solution, agents are asked to integrate nodes from the pool to reestablish a feasible solution and are not allowed to make any other changes to their local states.
We note that the local agent actions which involve local re-ordering or giving nodes to the pool cannot interfere with each other. To guarantee the same for taking nodes from the pool, the pool has a coordinating mechanism which offers nodes to agents for integration in a conflict-free manner (cf.\ Section \ref{models}).
The designed pool component thus enables us to meet the principles of conflict-free decisions and also self-determined agents, since all communication regarding exchanging nodes flows through the pool and nodes from there are integrated on a voluntarily basis.
The rules of the game, including the correspondence of the described rewriting actions to the choices of region and rule, are formalized in Appendix \ref{app_game}. 
\newline\newline
\textbf{Rewards}\hspace{0.2cm}
The global reward generally reflects the improvement in the team average cost for two feasible global states and penalizes the agent team if infeasible solutions were created consecutively for a too long time.
The team average cost in the global state $\displaystyle \vs_t$ is given by $c(\vs_t) = \frac{1}{n} \sum_i^n \evc^i(\evs_t^i)$ where $\displaystyle \evc^i(\displaystyle\evs_t^i)$ denotes the local cost of agent $\displaystyle i$ at time $\displaystyle t$. 
The global reward at time $\displaystyle t$ is then defined by
\begin{equation}
   \displaystyle \evr_t = \begin{cases}
       \displaystyle c(\vs_{prev_f(t)})-c(\vs_t) & \text{if $\displaystyle \vs_t$ is feasible,}\\
      -10 & \text{if $\displaystyle \vs_t$ is infeasible and the last $\displaystyle m$ global states were infeasible,}\\
      0 & \text{else,}
    \end{cases}    
    \label{reward}
\end{equation}
where $\displaystyle c(\displaystyle\vs_{prev_f(t)})$ denotes the team average cost of the last feasible solution before time step $\displaystyle t$ and $\displaystyle m$ is a hyperparameter. A strictly positive reward thus indicates an improved corresponding current feasible solution compared to the last one. For $m > 1$ the reward becomes non-Markovian which is currently left to the RL algorithm to learn about instead of being explicitly handled in the state representation. The choice of $m$ is discussed in Appendix \ref{app_hyperparam}. Note that computing the global reward requires the agent-specific costs in a global state to be revealed during training. 
\newline\newline
\textbf{Rewriting episode}\hspace{0.2cm}
The globally observed episode starts with an initial feasible global state at time zero and is limited by a fixed number of rewriting steps $\displaystyle T$: $(\displaystyle\vs_0,\displaystyle\va_0,\displaystyle\evr_1,\displaystyle\vs_1,\displaystyle\va_1,\displaystyle\evr_2,\displaystyle\vs_2,...,\displaystyle\vs_{T-1},\displaystyle\va_{T-1},\displaystyle\evr_T, \displaystyle\vs_T)$.
The final solution to the routing problem is defined by the last feasible global state in the rewriting episode.

\subsection{Model Overview}
\label{models}
\begin{figure}[h]%
    \centering
    \subfloat[Training: Each agent samples multiple local candidate actions which are centrally collected to form global candidate actions. One of these global actions is centrally selected for execution.]{{\includegraphics[width=0.515\textwidth]{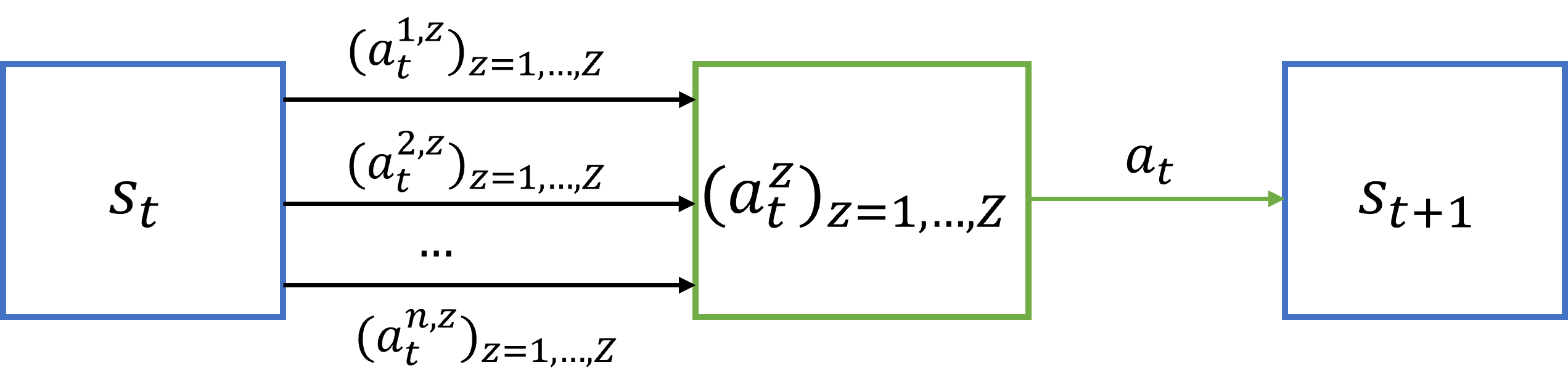} }}%
    \qquad\qquad
    \subfloat[Inference: Each agent chooses one local action which automatically determines the global action in a decentral manner.]{{\includegraphics[width=0.3\textwidth]{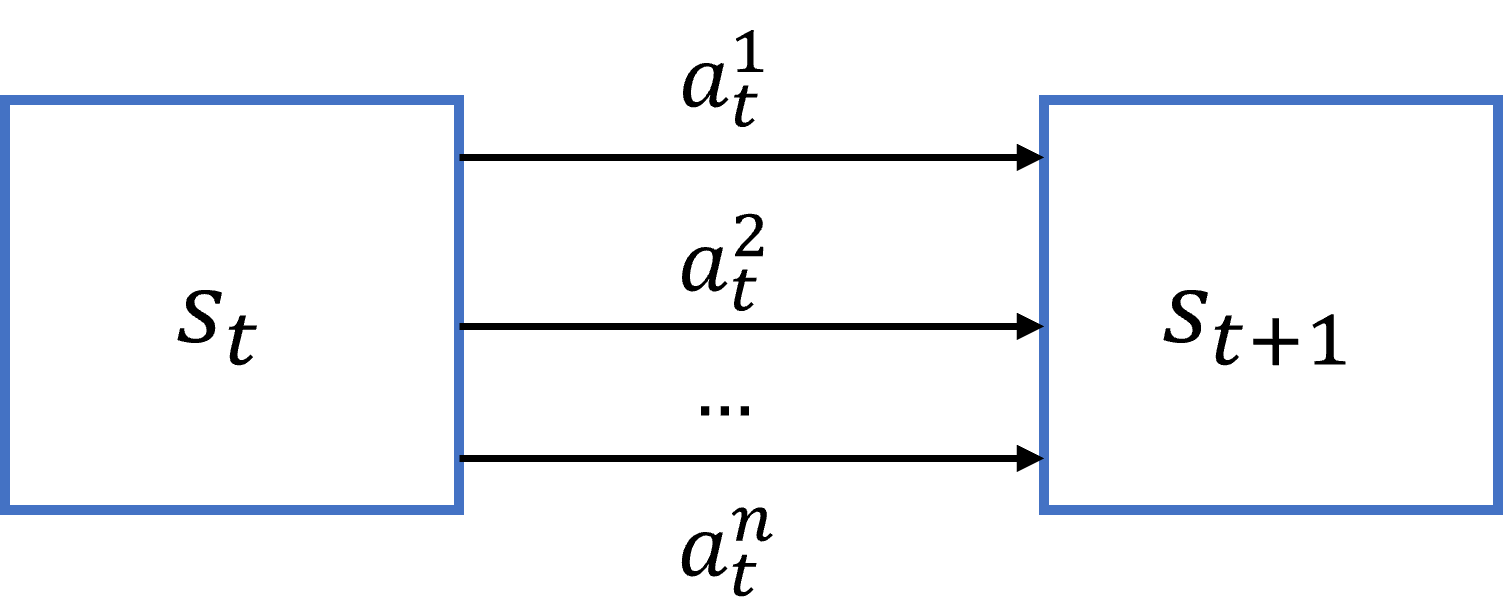} }}%
    \qquad
    \caption{Global action determination: training vs.\ inference.}
    \label{fig_global_action_training_inference}
\end{figure}

In this section, we present the RL-based models used to implement the described rewriting game.\newline
We saw that a local agent action is a two-step procedure, requiring a region and a corresponding rule node to be chosen. Defining one policy over the tuple of regions and rules would result in a discrete distribution with a sample space size that is quadratic in the problem size.
Hence we follow \citet{chen2019learning} and reduce the space by considering two separate distributions for regions and rules. In our current implementation, only the rule distribution is modelled as a neural network, the region distribution is random (while following some rules). It can harm the overall rewriting procedure only in terms of slowing it down, as described below in more detail.
Since we assume homogeneous and thus interchangeable agents, we learn a single agent-agnostic policy for choosing rules.
To ensure limited disclosure of local costs in the execution phase, we require the agent policy to only observe local cost information for decision-making. This is enough information for agents to help the team by optimizing the order of their own currently visited nodes. However, it is not sufficient for deciding whether exchanging nodes is beneficial for the team. Hence, we must provide them with global cost knowledge during learning. This is realized by an actor-critic approach following the scheme of centralized-learning-decentralized-execution: Based on perfect knowledge about all agent costs, the centralized critic learns to judge global actions in given global states with respect to their benefit for the team. The critic is used to guide the centralized agent policy. It enables the agent policy to learn about the underlying cost matrix distribution and thus to locally assess if integrating or excluding a node is helpful for the team. To facilitate the training process of the critic, we don't learn it by a posteriori showing it a global action defined by the chosen local agent actions, but let the critic centrally select the global action itself during training. It allows a better trade-off between exploration and exploitation and thus helps learning, especially given the high-dimensional joint action space. We force agents to centrally coordinate their local actions only during training. During inference, solely the agent policy is used in a decentral manner, see Figure \ref{fig_global_action_training_inference}.

In the following, we summarize the workflow for generating a global action during training and inference together with the necessary model components. Due to space limitations, we refer the reader to Appendix \ref{app_loss} for a discussion of the corresponding loss functions. 
\newline\newline
\underline{Encoding nodes:} For each node in the problem, we learn high-dimensional embeddings which incorporate information about the current state. Each agent encodes its currently visited nodes solely based on local information with an LSTM-based agent-agnostic local state encoder. For nodes in the pool we introduce an additional model, the pool state encoder, as the pool state differs from the local ones in its semantic structure. These learned embeddings are fed into our centralized critic and agent rule policy for decision-making. See Appendix \ref{app_models} for more details on the node encoders.
\newline\newline
\underline{Determining local actions:} During training, we sample $\displaystyle Z > 1$ candidate local actions for each agent. During inference, each agent selects one local action ($\displaystyle Z=1$) (cf.\ Figure \ref{fig_global_action_training_inference}).
A local action requires to choose a region node to be moved and based thereon a rule node. These decisions are made successively with two different models: 
A region node is determined with the region selector which is influenced by randomness. The corresponding rule node is chosen by a learned agent-agnostic but agent-cost dependent rule policy, the local rule selector.\newline
\textbf{Region selector}\hspace{0.2cm}
Region nodes are chosen randomly while following some rules. For local re-ordering or exclusion of a node, agents can independently sample from a uniform distribution defined over their currently visited customers to get candidate local region nodes. 
If the pool is filled and a region node must thus be one of the customer nodes in pool, we coordinate the assignment of region nodes to agents to avoid conflicts. The coordination can be viewed as an automated mechanism of the pool which simply informs each agent about its (for it) selected region.
This mechanism is also driven by randomness but at the same time equipped with a little intelligence. It neither offers a node first to the agent who just dropped the node in the pool, nor asks the same agent to integrate a node multiple times in a row (throughout the rewriting episode).
We note that a random region selector does not necessarily aggravate the rewriting procedure in terms of leading to a bad rewritten state since an agent can always choose to do nothing via the rule node if a bad region node was selected. Nevertheless it can slow down the rewriting procedure.\newline
\textbf{Local rule selector}\hspace{0.2cm} Given an agent's region node, the learned agent-agnostic local rule selector completes a local action by selecting a corresponding rule node. The decision is based on a predicted probability distribution over all possible rule candidates. It relies on an attention mechanism which processes node encodings of the already selected region and all respective possible rules as well as information as of how choosing the rule would affect the agent's local state (in terms of its node representations).
During training, the rule is sampled from the predicted probability distribution while for inference we choose the rule with the highest probability.
\newline\newline
\underline{Determining a global action:} During training, the $\displaystyle Z$ candidate local actions per agent are centrally collected and zipped together to build $\displaystyle Z$ candidate global actions. We choose one global action out of these candidates for rewriting based on an over the joint action space learned action-value function. 
During inference, the single chosen local agent actions automatically determine the global action (cf. Figure \ref{fig_global_action_training_inference}) . \newline
\textbf{Global action scorer}\hspace{0.2cm} The learned MLP-based global action scorer quantifies the expected team benefit of rewriting a given global state with a given global action. During training, we use it in an epsilon-greedy strategy to establish a good trade-off between exploration and exploitation.

\section{Empirical Evaluation}
\label{sec_emp_eval}
We empirically evaluate the MANR for vehicle routing in different setups on simulated data varying the number of agents as well as the problem size, i.e., the number of customer nodes in the routing problem. The procedure for data simulation is summarized in Section \ref{sec_data_gen}.
Section \ref{sec_exps} discusses the experiment results for all setups and compares them to a benchmark. There exists no comparable (RL-based) approach which could be used in the same collaborative limited disclosure setting out of the box. For this reason, we compare our results to an established approach with perfect cost knowledge and evaluate the competitiveness of our method. We chose the widely used OR-Tools optimization software as a benchmark which is also based on the concept of local search. In Appendix \ref{app_collaboration} we furthermore demonstrate the benefit of collaboration for the average agent in our approach by comparing it with a non-collaborative setup.

\subsection{Data Generation}
\label{sec_data_gen}
\begin{wrapfigure}{r}{7.5cm}
    \begin{tabular}{@{}cc@{}}
    \hspace{-9pt}
    \includegraphics[width=0.31\textwidth]{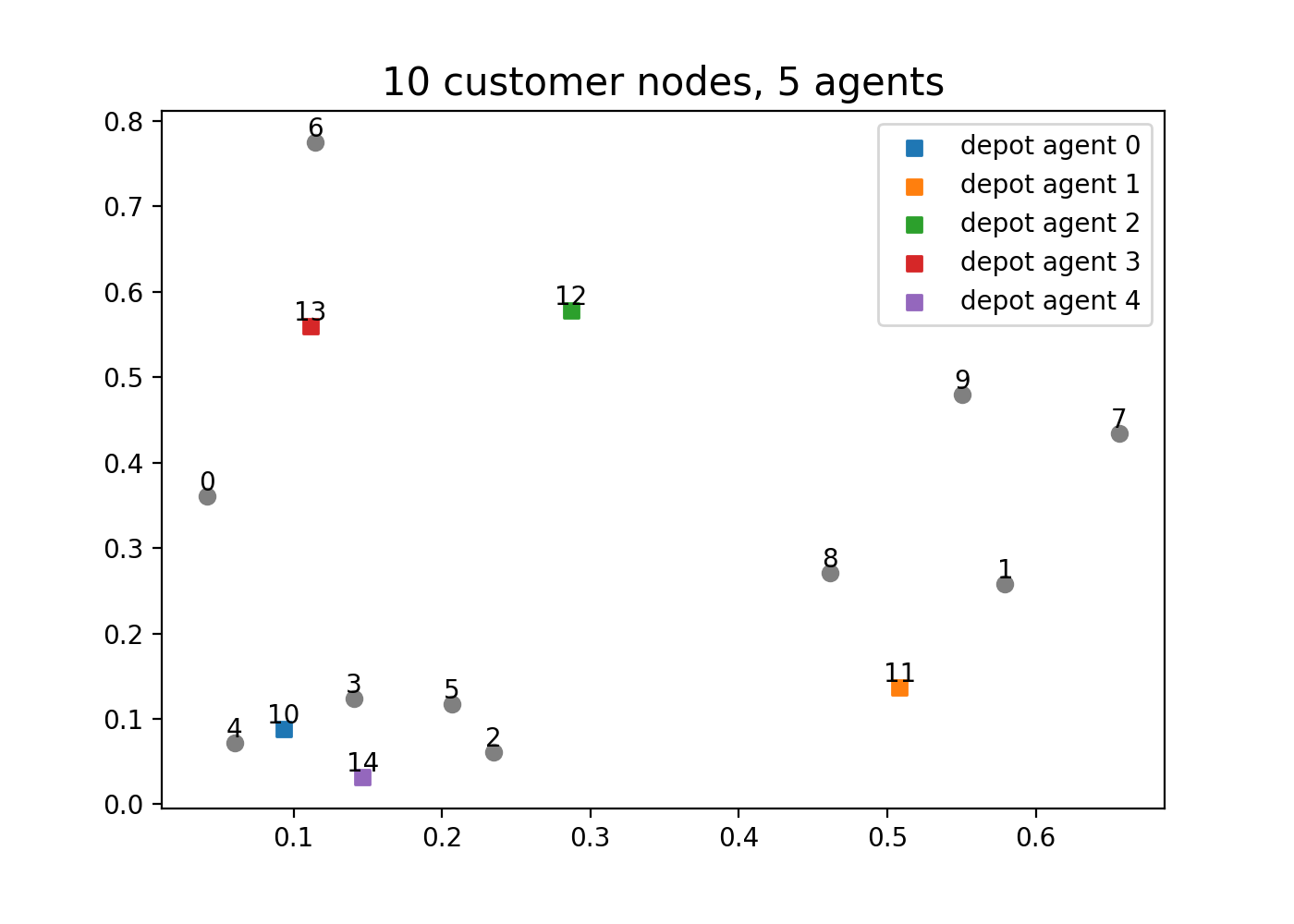} &
    \hspace{-34pt}
    \includegraphics[width=0.31\textwidth]{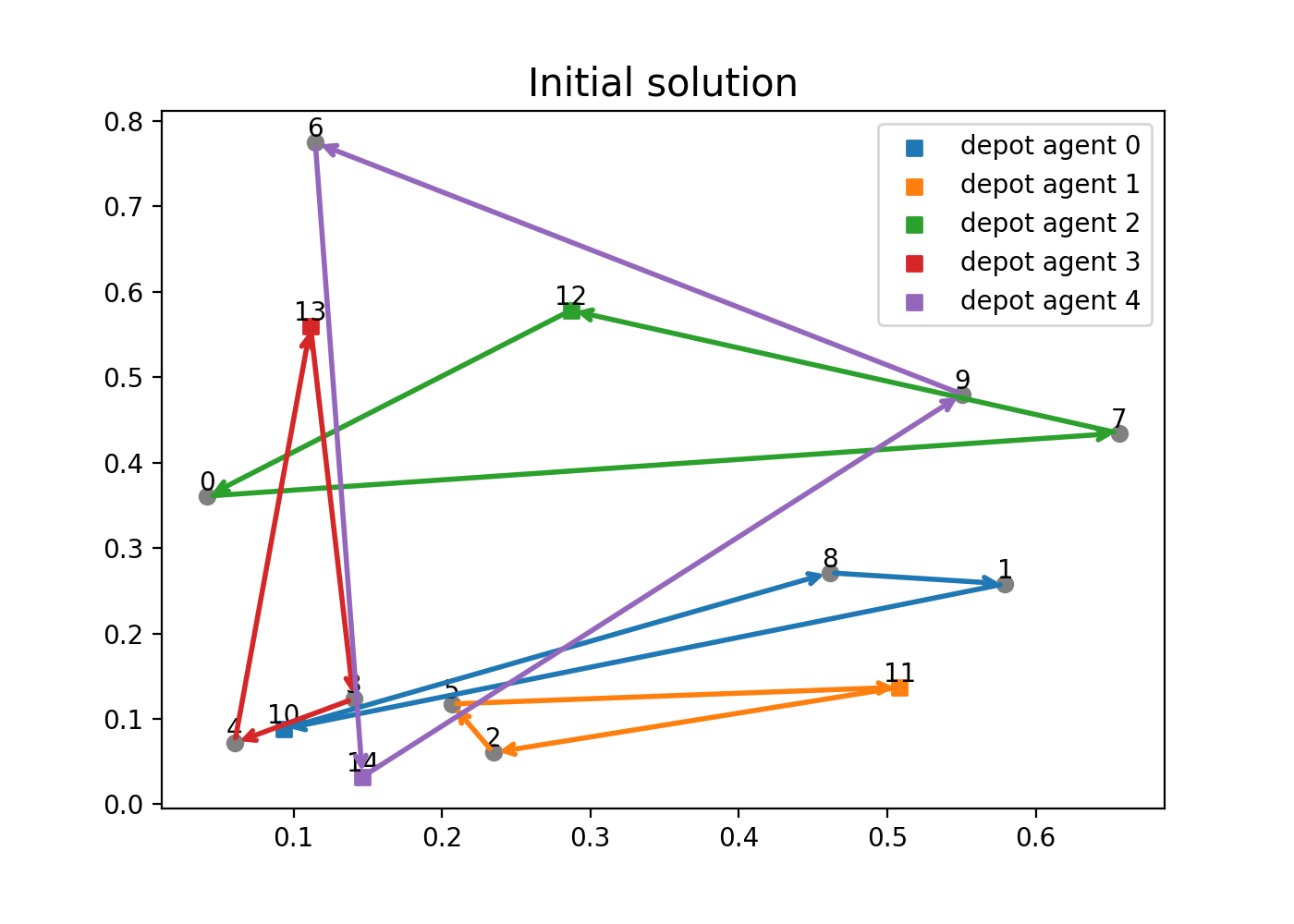}
    \end{tabular}
    \caption{Sampled routing problem with a corresponding sampled initial solution for $10$ customers and $5$ agents.}
    \label{fig_data_init}
\end{wrapfigure}

Customer nodes and agent depot nodes are sampled within the unit square, see Figure \ref{fig_data_init}. We draw a random fraction of customer nodes near the agent depots to enforce the participation of all agents in an optimal routing solution. The precise node sampling procedure is explained in Appendix \ref{app_datagen}.
The agent-specific costs between two customer nodes are modelled via agent-specific velocities. Each agent velocity $\displaystyle \eta^i$ is uniformly sampled from $[0.95,1]$, i.e., an agent can be at most $5\%$ faster than other agents. The cost for agent $\displaystyle i$ to travel between two nodes $\vv, \vz \in [0,1]^2$ is then given by their inverse-velocity-scaled Euclidean Distance as $\displaystyle \evc^i(\vv,\vz) = \tfrac{1}{\eta^i} \lVert \vv-\vz \rVert_2 $.
The MANR requires an initial feasible solution as a starting point for the rewriting procedure. To enforce the pool usage need for successful rewriting, we simply randomly assign the customer nodes to agents as evenly as possible. Each agent then applies the nearest neighbour heuristic \citep{rosenkrantz1977analysis} to its customer node set, see Figure \ref{fig_data_init} for an exemplary initial solution. Further examples of initial states are depicted in Appendix \ref{app_datagen}.

\subsection{Experiments}
\label{sec_exps}
We performed experiments for vehicle routing problems of sizes $10$ and $20$ with $2$,$3$ and $5$ agents respectively. Each data set consists of $6280$ vehicle routing problems and is split into three parts of $80\%$-$10\%$-$10\%$ for training, validating and testing. Hyperparameter tuning was performed on the validation set with Ray Tune\footnote{https://docs.ray.io/en/latest/tune/}. During training, we chose $30$ rewriting steps in all setups with $10$ customer nodes and $40$ rewriting steps for those with $20$ nodes. For inference, we increased the rewriting steps to compensate the stochastic region selection: In contrast to the training phase, each agent now just samples a single random region in a step; we don't have the luxury of choosing from a set of candidate actions with (most probably) different random region suggestions. We consider $100$ rewriting steps throughout all experiments for evaluation which we have found to work out well for all setups. We refer to Appendix \ref{app_hyperparam} for a complete overview of selected hyperparameter values for each of the experiments. We compare our results to the routing solver from OR-Tools which is based on local search and specifically tuned for vehicle routing\footnote{https://developers.google.com/optimization/routing/vrp}. We use the default search parameters and start with the same initial solutions as the MANR. Note that the solver necessitates complete knowledge about all vehicle costs for optimization.
\newline\newline
\textbf{Evaluation}\hspace{0.2cm} For each experiment, we make $\displaystyle 20$ inference runs on the test set (due to the region stochasticity) and compute the test set performance averaged over all runs. Performance is measured with the team average cost of the last feasible solution in the rewriting episode and the initial solutions are kept equal throughout all runs. We also compute the mean performance when choosing the best run for each test sample individually ("MANR best"). This is a natural and valid metric when inference time is not decisive in the application. We compare these values to the mean test set performance of the initial solutions as well as of the solutions produced by OR-Tools: We report the performance gaps in terms of percentage cost reductions relative to the intitial solution and relative to OR-Tools. We also report the average run times for one routing problem, when performing evaluation on a server with a single GPU (Tesla V100S-PCIE-32GB) and CPU core. The results for the setup of $10$ nodes are presented in Table \ref{table_eval_10}. The results for $20$ nodes as well as the absolute performance values for all setups can be found in Appendix \ref{app_eval_figures}.

\begin{table}[!htb]
    \caption{Empirical results for $10$ customer nodes and a varying amount of agents.}
    \label{table_eval_10}
    \begin{subtable}{0.545\linewidth}
      \centering
        \caption{Performance gaps} 
        \begin{tabular}{|c||c|c|c|c|}
        \hline
        \multirow{2}{*}{Setup} & \multicolumn{2}{c|}{gap init} & \multicolumn{2}{c|}{gap OR-Tools}  \\
        \cline{2-5}
        & MANR  & MANR best  & MANR & MANR best \\ \hline
        2 agents & 32\% & 40\% & -21\% & -7\%  \\ 
        3 agents & 41\% & 51\% & -37\% & -14\%\\ 
        5 agents & 51\% & 62\% & -59\% & -24\%\\ \hline
        \end{tabular}%
    \end{subtable}
    \begin{subtable}{0.545\linewidth}
      \centering
        \caption{Average run time in seconds} 
        \begin{tabular}{|c|| c| c|}
            \hline
          \multirow{2}{*}{Setup} & \multirow{2}{*}{MANR} & \multirow{2}{*}{OR-Tools} \\
          &&\\
             \hline
            2 agents & 0.31  &  0.01 \\  
            3 agents  &  0.47  & 0.01  \\ 
            5 agents & 0.71 & 0.01 \\  
            \hline
        \end{tabular}
    \end{subtable}%
\end{table}

\textbf{Discussion}\hspace{0.2cm}
For all experimental setups (both $10$ and $20$ nodes), the MANR significantly improves over the initial solutions: We have an average percentage cost reduction of $40\%$ for the mean MANR solution over all runs ("MANR") and $50\%$ when considering the best solution for each routing problem per run ("MANR best"). Generally, the more agents the bigger the percentage cost reduction. This stems from the fact that the more agents, the higher the probability that a node gets assigned to a wrong agent in the randomly generated initial solution. Hence, we expect worse initial solutions and thus more room for improvement with an increasing amount of agents. The performance of the OR-Tools benchmark cannot be reached by the MANR but it gets reasonably close, taking into account the imperfect cost knowledge of the MANR agents. Evaluating the results of both $10$ and $20$ nodes, there is an average percentage cost reduction of $-41\%$ of the MANR solution quality relative to the one of OR-Tools, respectively, $-14\%$ when considering MANR best. The more agents the more pronounced the gap. This can be explained with the increasing global action space dimensionality which makes it more difficult for the global action scorer (the critic) to learn.
Looking at individual rewriting episodes of the MANR, we saw that agents learned to use the newly introduced pool in a meaningful way (see Appendix \ref{app_rollout} for an excerpt of an exemplary rewriting rollout). We also observed, that the MANR even solves a few isolated problems better than OR-Tools. It will be interesting to analyze the kind of these routing problems in the future. Comparing inference times, the well-tuned, in C++-implemented OR-Tools software is significantly faster. It is also less influenced by the number of agents. However, it requires to be perfectly informed about all agent costs in contrast to the MANR. Also, up to now, we haven't optimized our Python code for efficiency. There is room for improvement, e.g., by parallelizing the decentral agent action generation during inference which currently runs sequentially in our implementation. We could also investigate in more detail, which number of rewriting steps suffices in the respective settings and save time with shorter rewriting rollouts. Moreover, we expect the run times to approach each other when scaling to larger problem sizes as it was observed for the Neural Rewriter.

\section{Conclusion}
\label{conclusion}
We presented the multi-agent Neural Rewriter (MANR) for collaboratively solving a multi-vehicle routing problem with limited disclosure of vehicle-specific costs in spirit of a team Markov game. Vehicle-specific costs are only explicitly shared in the training phase to allow agents to learn about the underlying cost distribution. During inference, each agent performs its action solely based on local cost information.
We enable parallel conflict-free agent actions in the game by introducing a pool mechanism which coordinates the necessary node exchange between agents. The introduced pool comes at the cost of generating also infeasible solutions within the rewriting episode, but is counteracted by teaching agents a proper pool usage during training.
Our agent-agnostic policy, which must solely process local information, is provided with some global knowledge during training via its cost-omniscient critic.
Our empirical results demonstrate that the approach indeed enables the solely local cost observing agents to act for the sake of the team, i.e., to exchange nodes with other agents via the pool in a meaningful way: The MANR improves an initial solution where nodes get simply randomly assigned to agents by $50\%$ on average. Agents learn to assess the capabilities of representative team members and can base their rewriting decisions thereon. 
The experiments also confirm the inescapable trade-off between non-disclosure and optimality. The performance of our benchmark, the OR-Tools heuristic, which assumes perfect cost information cannot be reached on average. However, the MANR gets close and even solves a few isolated problems better. 

In the future, we plan to further empirically evaluate the scalability of our approach by increasing the number of nodes in the routing problem. We want to improve our current implementation by fine-tuning the model architectures and enhancing the code efficiency as we expect a significant performance boost from it. Another adaption of the current setup could involve explicitly representing the pool history in the state by, e.g., introducing a counter counting the consecutive steps of a filled pool, to guarantee the Markov property in the game setup and thus simplify learning. The approach could furthermore be improved by replacing the random component in the region selection with a learned model. Also the issue of limited agent scalability due to joint action space learning needs to be addressed. Moreover, we plan to extend to other setups. One direction could be to consider a heterogeneous agent team where the agent cost matrices are not sampled from the same distribution. It requires to provide the agent policy with more global cost knowledge than in the current setup and hence intensifies the trade-off between non-disclosure and optimality.
Another interesting direction concerns to tighten the limited disclosure requirement by also demanding it during the learning phase. A possible way to avoid explicit revelation of agent-specific costs during training is to employ distributed machine learning approaches, i.e., to only share local model parameters. We could also let agents exchange abstract information via learned vector-based embeddings. For both options it would be interesting trying to quantify the non-disclosure: We could introduce an opponent to the game which tries to re-construct agent costs from the shared information.   

\subsubsection*{Acknowledgments}
The research of N.\ Paul was supported by the Fraunhofer Society within the project ``SWAP – Hierarchical swarms as production architecture with optimized utilization''. The work of T.\ Wirtz was funded by the German Federal Ministry of Education and Research, ML2R - no. 01S18038B. S.\ Wrobel contributed as part of the University of Bonn and the Fraunhofer Center for Machine Learning within the Fraunhofer Cluster for Cognitive Internet Technologies.

\bibliography{gmas_iclr2022_conference}
\bibliographystyle{gmas_iclr2022_conference}

\newpage

\appendix
\section{Appendix}
\label{appendix}
\subsection{Differences to the original Neural Rewriter}
\label{app_diffNR}
We give an overview about the most essential differences of the MANR compared to the original Neural Rewriter.
\newline\newline
\textbf{Application}\hspace{0.2cm} The Neural Rewriter is used for solving a capacitated vehicle routing. We don't consider the setup of vehicle capacities and customer demands and thus removed all associated information, e.g., in the node representations.
\newline\newline
\textbf{Game Setup}\hspace{0.2cm} The MANR extends the setup of the Neural Rewriter by considering multiple vehicles instead of a single one. As described in Section \ref{MANR}, it required a redefinition of the RL setup, which goes beyond simply scaling from local states and actions to global states and actions.
A key difference is the introduction of the pool set which is modelled as an additional component in the MANR's global state.
For the routing application, the Neural Rewriter technically allows an action to swap two nodes in the route. We use a different ruleset which is also mentioned by the authors \citet{chen2019learning}, that allows an action to move one node to a different position in the route. This is essential for the MANR as swapping two nodes would condemn an agent to always stay with the same amount of visited nodes and never allow it to reduce or enrich its visited node set.
At last, we adapt the reward definition of the Neural Rewriter to include feedback to proper and improper pool usage.
\newline\newline
\textbf{Models}\hspace{0.2cm} We follow the approach of the Neural Rewriter to consider separate models for the region and rule selection, but realize them differently. Moreover, in contrast to the Neural Rewriter, we learn a model to evaluate a complete (joint) action.\newline
The Neural Rewriter selects a region via a learned action-value function, while we currently model it with a rule-based random component. We instead learn an action-value function for the joint global action as our approach relies on an entity with perfect global knowledge. The action-value function of the Neural Rewriter learns to predict the expected (discounted) \textit{maximal} cost improvement compared to the given global state when choosing a region in the global state. This is meaningful taking into account that one is only interested in the best solution which was produced within the rewriting episode.
This idea is however not transferable to our setup with limited cost disclosure: In order to determine the best team solution in a rolled out rewriting episode, each agent would need to share its occuring costs for all produced solutions. For this reason, we instead always take the last feasible solution in the episode as the final solution, without evaluating any team average costs of solutions. We hence adapt the loss function for our action-value function and let it predict the (discounted) \textit{cumulative} team cost improvement.\newline
Our rule policy is almost identical to the one of the Neural Rewriter, we only provide it with (partly) different input: To decide which node is chosen as the rule, some fictitious node representations in the spirit of "what-would-happen-if" are provided (see definition of node representations within the local state and pool encoder in Appendix \ref{app_models}). For the Neural Rewriter they create a fictitious node representation for the candidate rule node after being swapped with the region and for another self-defined node (which is not part of the real nodes). 
In the MANR we create such fictitious node representations for each node whose representation would be affected by the local action, i.e., each node whose predecessor would change since node representations contain predecessor information. Consequently, for the employed local action type of placing the region node after the rule, we create fictitious node representations for the region node, the region node successor and the rule successor.
Analogously to the Neural Rewriter, we train the rule policy with an actor-critic approach but replace the critic with our action-value function which assesses global actions.\newline
At last, for encoding nodes in the local state, we take over the LSTM-based encoder of the Neural Rewriter. For encoding nodes in the pool we introduce a new attention-based encoder. 

\subsection{More details on the rules of the game}
\label{app_game}
We formalize the set of allowed local agent actions which depend on the pool state.\newline

\newpage
Notation:
\begin{itemize}
    \item $\displaystyle \sV$ denotes the set of customer nodes,
    \item $\displaystyle \vs_t = (\displaystyle \evs^1_t, \displaystyle \evs^2_t, ...,\displaystyle \evs^n_t,\sP_t)$ denotes the global state at time $\displaystyle t$, consisting of local agent states $s_t^i$ at time $\displaystyle t$ and the corresponding current state of the pool $\displaystyle \sP_t$.
\end{itemize}

\textbf{Local actions}\hspace{0.2cm} A local action of agent $i$ generally consists of selecting a region node $\displaystyle \evw^i$ and afterwards a corresponding rule node $\displaystyle \evu^i$ with $\displaystyle \eva^i = (\displaystyle \evw^i, \displaystyle \evu^i)$ implying the region node to be moved by being placed after the rule node. For the interaction with the pool we introduce a representative pool node denoted by $\displaystyle p$.
The allowed local actions for agent $\displaystyle i$ at time $\displaystyle t$ are defined as follows:

\begin{itemize}
    \item If the pool $\displaystyle \sP_t$ is empty: The region node to be moved can be any customer node from the agent's own local state $\displaystyle \evw^i_t \in \sV_{|s_t^i}$. The fixed rule set $\displaystyle \evu^i_t \in \{s_t^i\}\cup \{p\}$ then semantically translates to either keeping its local state unchanged by choosing the current region predecessor in its local state as the rule, re-arranging the region node locally by choosing any other node from its local state as the rule or excluding the region node from its local state by choosing the pool node as the rule.
    \item If the pool $\displaystyle \sP_t$ is filled: The region node to be moved can be any customer node from the pool state $\displaystyle \evw^i_t \in \sP_t$. The fixed rule set $\displaystyle \evu^i_t \in \{s_t^i\}\cup \{p\}$ then semantically translates to either accepting an offer to integrate the region node locally by choosing any node from its local state as the rule or declining an offer and thus leave its local state unchanged by choosing the pool node as the rule.
\end{itemize}

\subsection{Training details}
\label{app_loss}
All presented models (local state encoder, pool state encoder, local rule selector, global action scorer) are trained centrally and simultaneously with a combined loss function:

\begin{equation}
    L(\theta, \phi, \psi) = L_a(\theta,\psi) + \alpha \, L_u(\phi,\psi) ,
    \label{total_loss}
\end{equation}
where 
\begin{itemize}
    \item $\displaystyle \theta$ denote the parameters of the global action scorer $\displaystyle Q$ with corresponding loss $\displaystyle L_a(\cdot)$,
    \item $\displaystyle \phi$ denote the parameters of the local rule selector $\displaystyle \pi_u$ with corresponding loss $\displaystyle L_u(\cdot)$,
    \item $\displaystyle \psi=[\psi_1, \psi_2]$ denote the parameters of the local state and pool state encoder, which are implicitly trained with both losses $\displaystyle L_a(\cdot)$ and $\displaystyle L_u(\cdot)$.
\end{itemize}

The global action scorer $\displaystyle Q$ is fitted to the cumulative discounted observed reward within the rewriting episode as
\begin{equation*}
  L_a(\theta,\psi) = \frac{1}{T} \sum_{t=0}^{T-1}\bigg( \sum_{t'= t}^{T-1}\gamma^{t'-t}\evr_{t'+1} -Q(\evs_t,\eva_t;\theta,\psi)\bigg)^2,
\end{equation*}
where $\displaystyle \gamma < 1$ denotes the discount factor.
The agent-agnostic local rule selector $\displaystyle \pi_u$ is learned by an actor-critic approach with the global action scorer $\displaystyle Q$ serving as the critic:

\begin{equation*}
    L_u(\phi,\psi) = \frac{1}{n}\sum_{i=1}^n \bigg(- \sum_{t=0}^{T-1} A(\evs_t,\evu_t^i, \eva_t^{-i}, \evw_t^i) \log \pi_u(\evu_t^i |\evw_t^i,\evs_t^i, \sP_t; \phi,\psi)\bigg),
\end{equation*} with 

 \begin{equation*}
     A(\evs_t,\evu_t^i, \eva_t^{-i}, \evw_t^i)
     = Q(\evs_t, \evu_t^i, \eva_t^{-i}, \evw_t^i; \theta,\psi) - \sum_{\substack{\text{all candidate}\\\text{rules } \Tilde{u}^i}} \pi(\Tilde{u}^i|\evw_t^i,\evs_t^i, \sP_t; \phi,\psi) Q(\evs_t,\Tilde{u}^i, \eva_t^{-i}, \evw_t^i; \theta,\psi),
 \end{equation*}
where $\displaystyle A(\evs_t,\evu_t^i, \eva_t^{-i}, \evw_t^i)$ denotes the advantage of choosing rule node $\displaystyle \evu_t^i$ in the global state $\displaystyle\evs_t$ given region node $\displaystyle \evw_t^i$ and all other agents' local actions $\displaystyle \eva_t^{-i} = (\eva_t^1, ..., \eva_t^{i-1}, \eva_t^{i+1}, ..., \eva_t^n)$.
 Note that it's arguably correct to criticize a local rule node in the context of all other agents' local actions since the choice of $\displaystyle a_t^{-i}$ does not effect the optimal action $a_t^i$ for agent $i$ under $Q$, see \ref{app_critic} below for a formalization of this statement. An agent can thus not be penalized or praised for his rule node's choice due to bad or good actions of the other agents. 

The hyperparameter $\displaystyle \alpha < 1$ in \eqref{total_loss} weakens the loss of the local rule selector to give its critic, the global action scorer, a head first to perform well. The total loss is minimized with the Adam optimizer.

\subsubsection{Property of the global critic \textit{Q}}
\label{app_critic}
Let $$\displaystyle Q^\pi(s,a) = \mathbb{E}_\pi \Big[  \sum_{t'= t}^{T-1}\gamma^{t'-t}\evr_{t'+1} | s_t=s, a_t=a\Big]$$ denote the expected return when choosing global action $a_t = (a_t^1, a_t^2, ..., a_t^n)$ in the global state $s_t$ and following the joint agent policy $\pi$.

Define $$\displaystyle f_s(a^{-i}) = \argmax_{a^i} Q^\pi(s, a^i, a^{-i})$$ as the local action $a^i$ of agent $i$ which maximizes $Q^\pi$ for given other agent actions 
$\eva^{-i} = (\eva^1, ..., \eva^{i-1}, \eva^{i+1}, ..., \eva^n)$ and the global state $s$.

Then it holds $$f_s(a^{-i}) = f_s(\widetilde{a^{-i}})$$ for any other agent actions $\widetilde{a^{-i}}$.
I.e., the optimal action of agent $i$ under $Q^\pi$ in a given global state $s$ is independent of the other agents' actions at the same time step.

\subsection{More details on the models}
\label{app_models}

\textbf{Local state encoder}\hspace{0.2cm} Each node $\displaystyle v$ in the local state of vehicle agent $\displaystyle i$ is represented by a 5-dimensional vector $\displaystyle (\evv_x,\evv_y, \evv^p_x, \evv^p_y, \evc^i(v, \evv^p))$ containing x,y coordinates of the node itself and of its current predecessor in the route $\displaystyle \evv^p$ as well as the agent-specific cost to travel between them. Each such node representation is encoded by a learned bidirectional LSTM which follows the trajectory of the local agent route and incorporates the sequential information.
\newline\newline
\textbf{Pool state encoder}\hspace{0.2cm} Each node $\displaystyle v$ in the pool state is represented by a 2-dimensional vector $\displaystyle(\evv_x,\evv_y)$ containing its x,y coordinates. Since there exists no semantic sequential order in the pool state, each such node representation is encoded by a learned self-attention mechanism.

\subsection{Collaboration benefit}
\label{app_collaboration}

In the following, we empirically demonstrate the incentive for the average agent to collaborate on serving customers:
Assume each agent has its own set of customers to be visited. For each agent separately, we solve a travelling salesman problem (TSP) computing an optimized route visiting all agent's customer nodes. The non-collaborative baseline cost $c_{\text{non-collab}}$ is then given by the average cost over all agent TSP solutions.
We compare the result to our collaborative approach where agents combine their customer orders to find an optimized assignment of customers to agents as well as the corresponding routes. Analogously to Section \ref{sec_exps}, we compute the team average cost for our setup, i.e., the cost for the average agent when collaborating, denoted by $c_{\text{team}}$ (both for "MANR" and "MANR best").
The collaboration benefit is then defined by the percentage cost reduction relative to the non-collaborative baseline cost as
$\tfrac{c_{\text{non-collab}}-c_{\text{team}}}{c_{\text{non-collab}}}$.

We report the collaboration benefit for all in Section \ref{sec_exps} discussed experimental setups of 10 and 20 customer nodes together with 2,3,5 agents respectively on the test set. For the non-collaborative baseline we consider the same agent-customer-assignments as in the initial solutions of the MANR (cf. Section \ref{sec_data_gen}). The resulting agent routes thus only differ from those in the initial solutions in such a way that they are not obtained by the nearest neighbour heuristic but are optimized with the TSP-solver from OR-Tools\footnote{https://developers.google.com/optimization/routing/tsp}.

Table \ref{table_app_collab} summarizes the empirical results. For $10$ customer nodes we have an average collaboration benefit of $40\%$ for MANR, resp. $50\%$ for MANR best. For $20$ customer nodes we have an average collaboration benefit of $33\%$ for MANR, resp. $45\%$ for MANR best. Generally, the more agents, the higher the profit for collaborating. This stems from the fact that the more (random) agent  depots the higher the chance that a different agent can visit a given customer node cheaper than the agent initially assigned to it.

\begin{table}[!htb]
    \caption{Empirical results for the collaboration benefit for a varying amount of customer nodes and agents.}
    \label{table_app_collab}
    \begin{subtable}{0.545\linewidth}
      \centering
        \caption{10 nodes} 
        \begin{tabular}{|c||c|c|}
        \hline
        \multirow{2}{*}{Setup} & \multicolumn{2}{c|}{collaboration benefit} \\
        \cline{2-3}
        & MANR  & MANR best  \\ \hline
        2 agents & 28\% & 37\% \\ 
        3 agents & 40\% & 50\% \\ 
        5 agents & 51\% & 62\% \\ \hline
        \end{tabular}%
    \end{subtable}
    \begin{subtable}{0.545\linewidth}
      \centering
        \caption{20 nodes} 
        \begin{tabular}{|c||c|c|}
        \hline
        \multirow{2}{*}{Setup} & \multicolumn{2}{c|}{collaboration benefit} \\
        \cline{2-3}
        & MANR  & MANR best  \\ \hline
        2 agents & 20\% & 31\% \\ 
        3 agents & 33\% & 47\% \\ 
        5 agents & 45\% & 58\% \\ \hline
        \end{tabular}%
    \end{subtable}
\end{table}

\subsection{Data generation}
\label{app_datagen}
\textbf{Nodes}\hspace{0.2cm} When sampling all customer nodes uniformly in the unit square as in \citet{chen2019learning}, we observed that it is often overall cheaper if only one agent sets off to do the job alone. To diversify the data set and enforce the participation of all agents in an optimal routing solution, we only draw a random fraction of customer nodes uniformly and the rest near the agents' depots. More precisely, we first sample agent depot node coordinates $\displaystyle \evd^i$, $i \in \{1,...,n\}$, uniformly in the unit square $[0,1]^2$. For each agent $\displaystyle i$ we define a bivariate truncated normal distribution in the unit square $\displaystyle\mathcal{N}(\evd^i,\,0.1^{2})$ centered at its depot with a standard deviation of $0.1$. 
We randomly sample the fraction of customer nodes which shall be drawn uniformly in the unit square. The remaining customer nodes are assigned to the agents as evenly as possible and their coordinates are drawn from the described agent-specific normal distributions.
\begin{figure}[!htb]
     \centering
    \begin{subfigure}[t]{1\textwidth}
        \raisebox{-\height}{\includegraphics[width=0.32\textwidth]{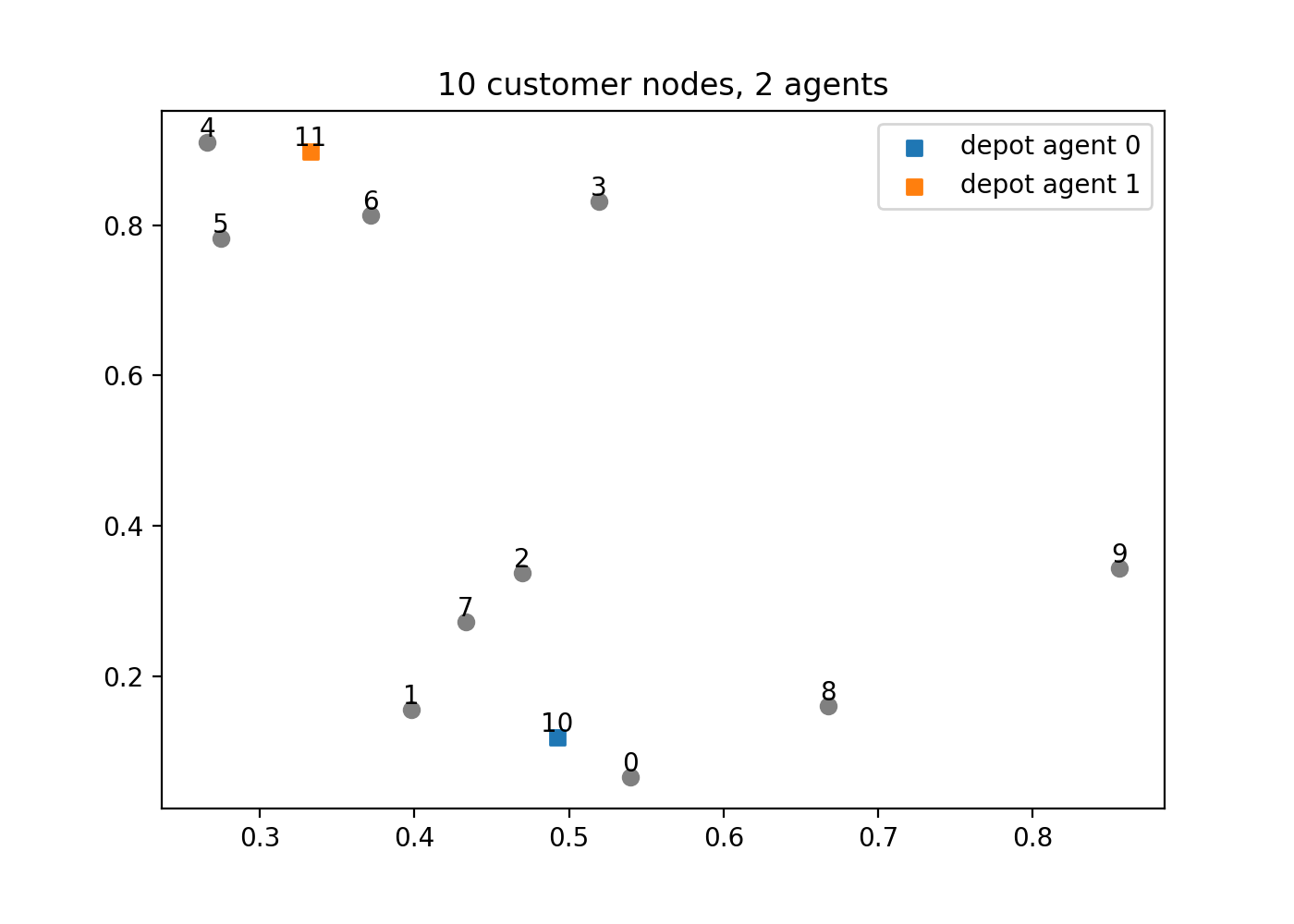}}%
        \raisebox{-\height}{\includegraphics[width=0.32\textwidth]{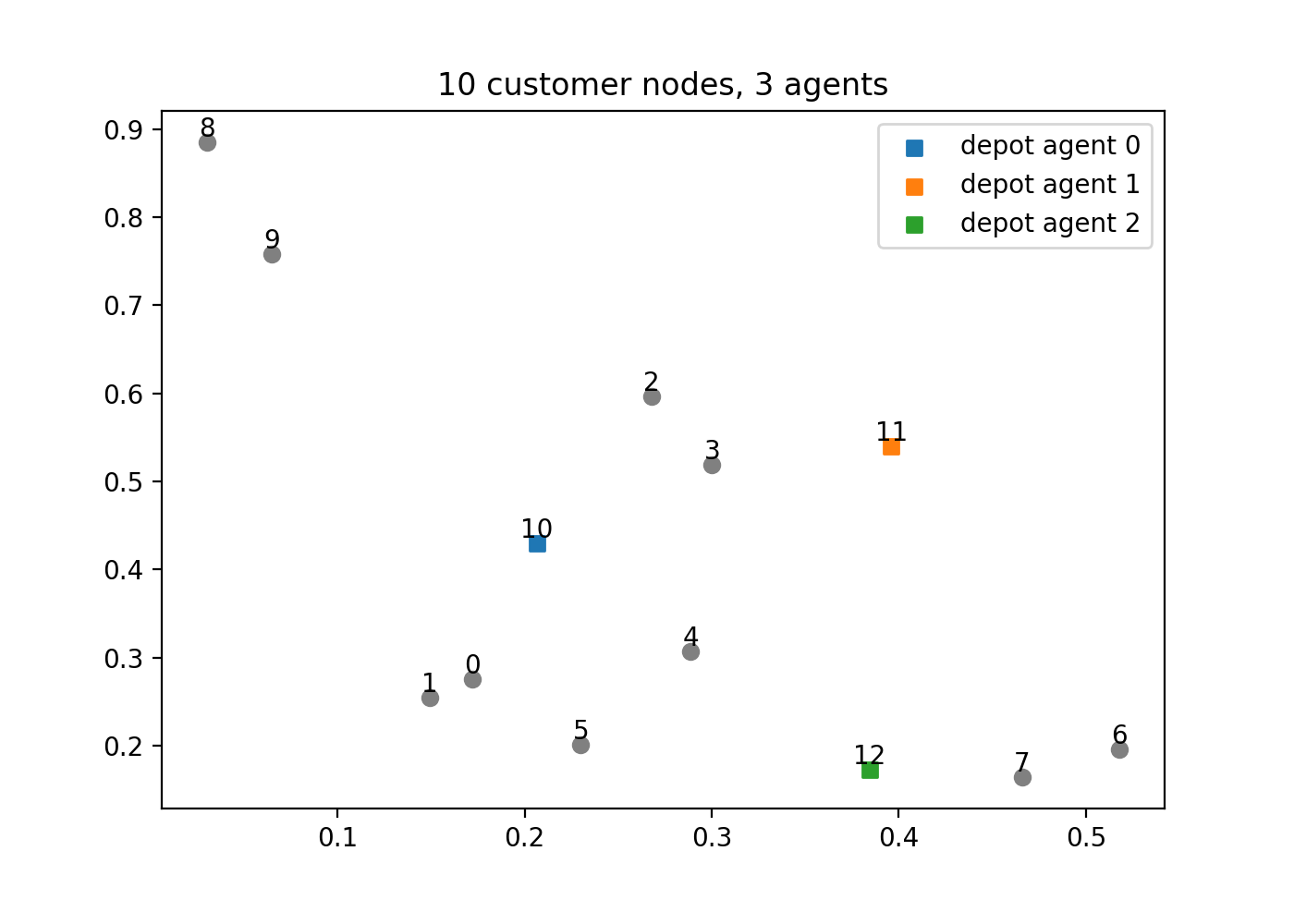}}%
        \raisebox{-\height}{\includegraphics[width=0.32\textwidth]{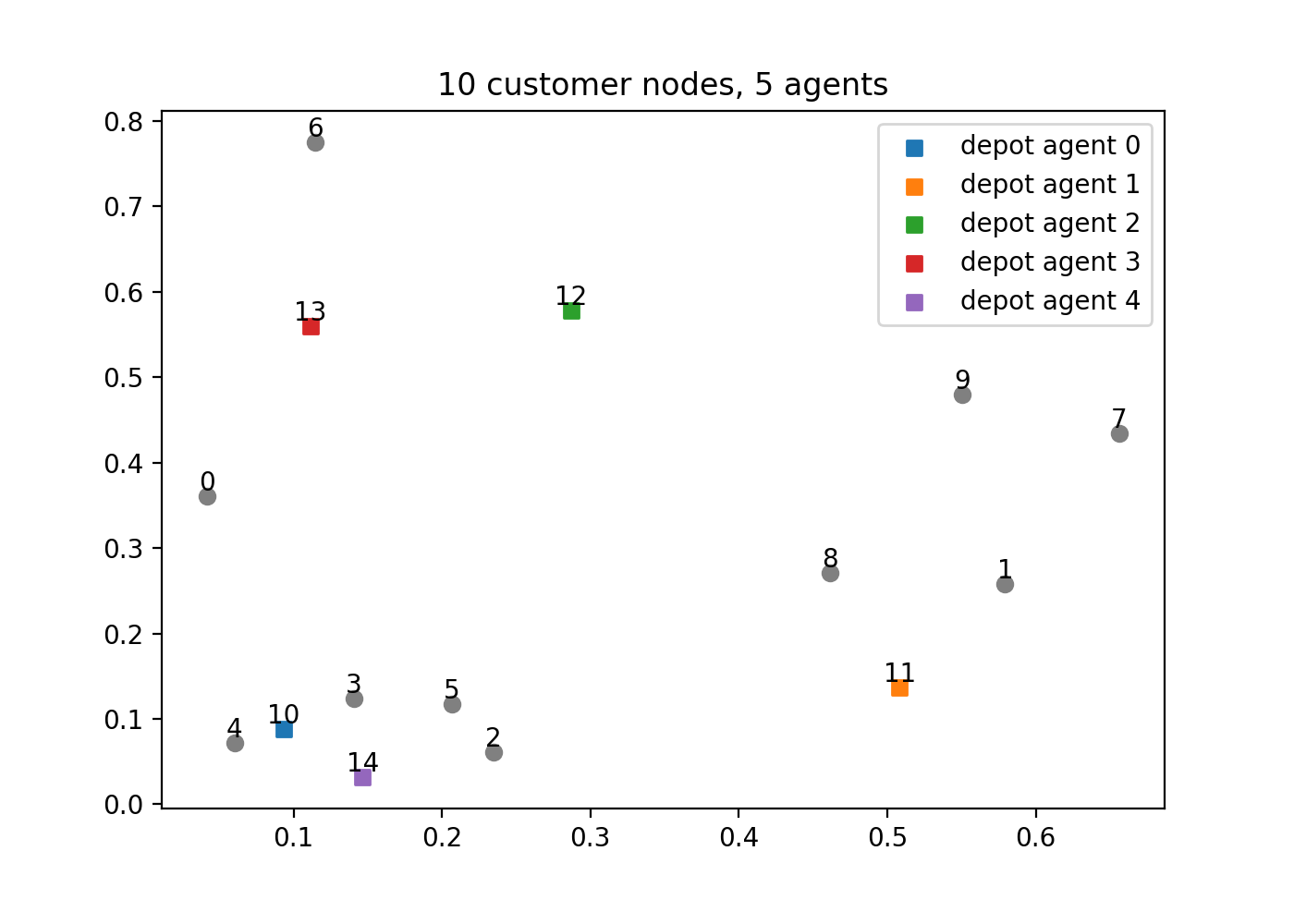}}%
    \end{subfigure}
    
    %%%%%%%%%%%%%%%%%%%%%%%%%%%%%%%%%%%% second row
    \begin{subfigure}[t]{1\textwidth}
        \vspace{-10pt}
        \raisebox{-\height}{\includegraphics[width=0.32\textwidth]{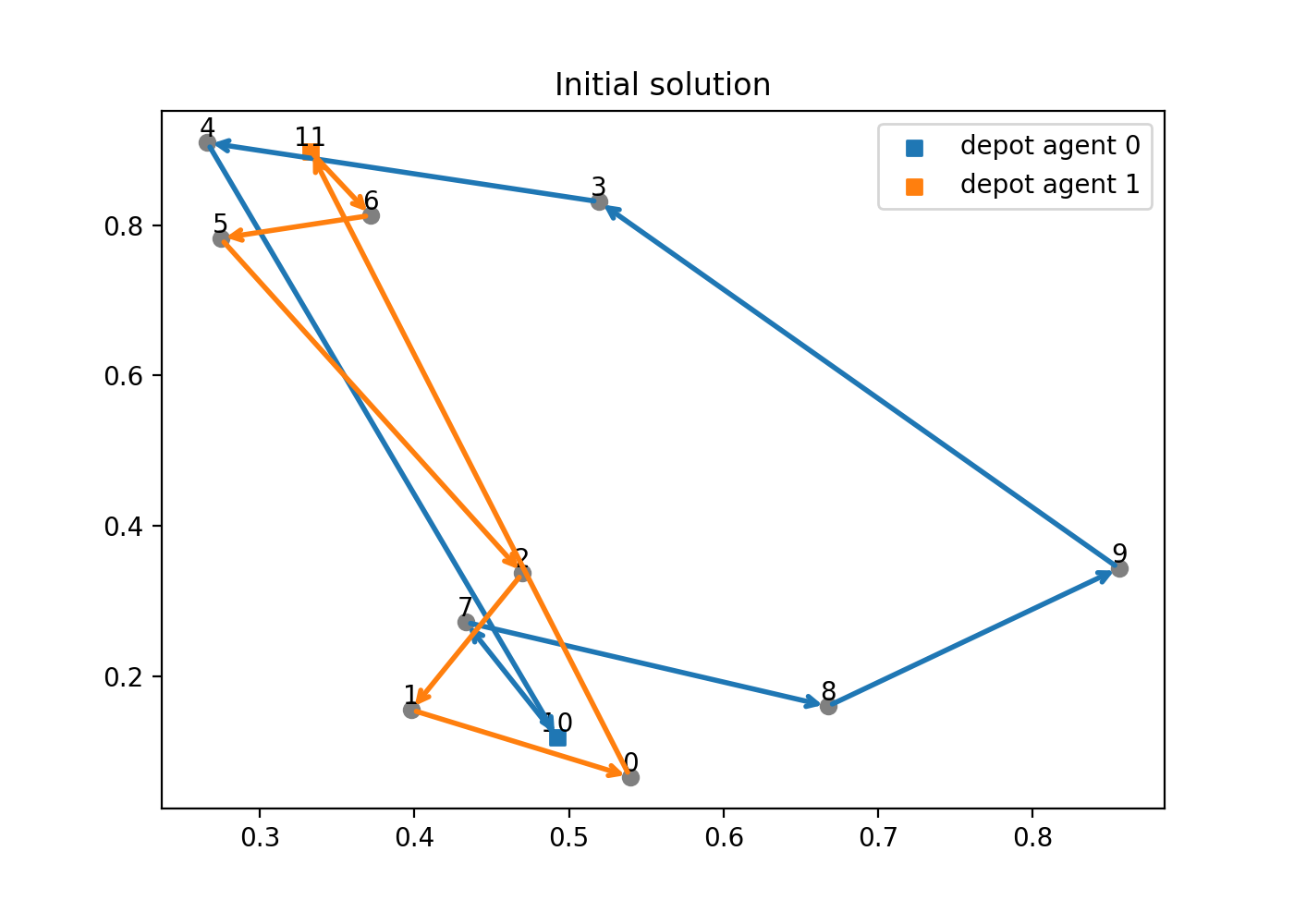}}%
        \raisebox{-\height}{\includegraphics[width=0.32\textwidth]{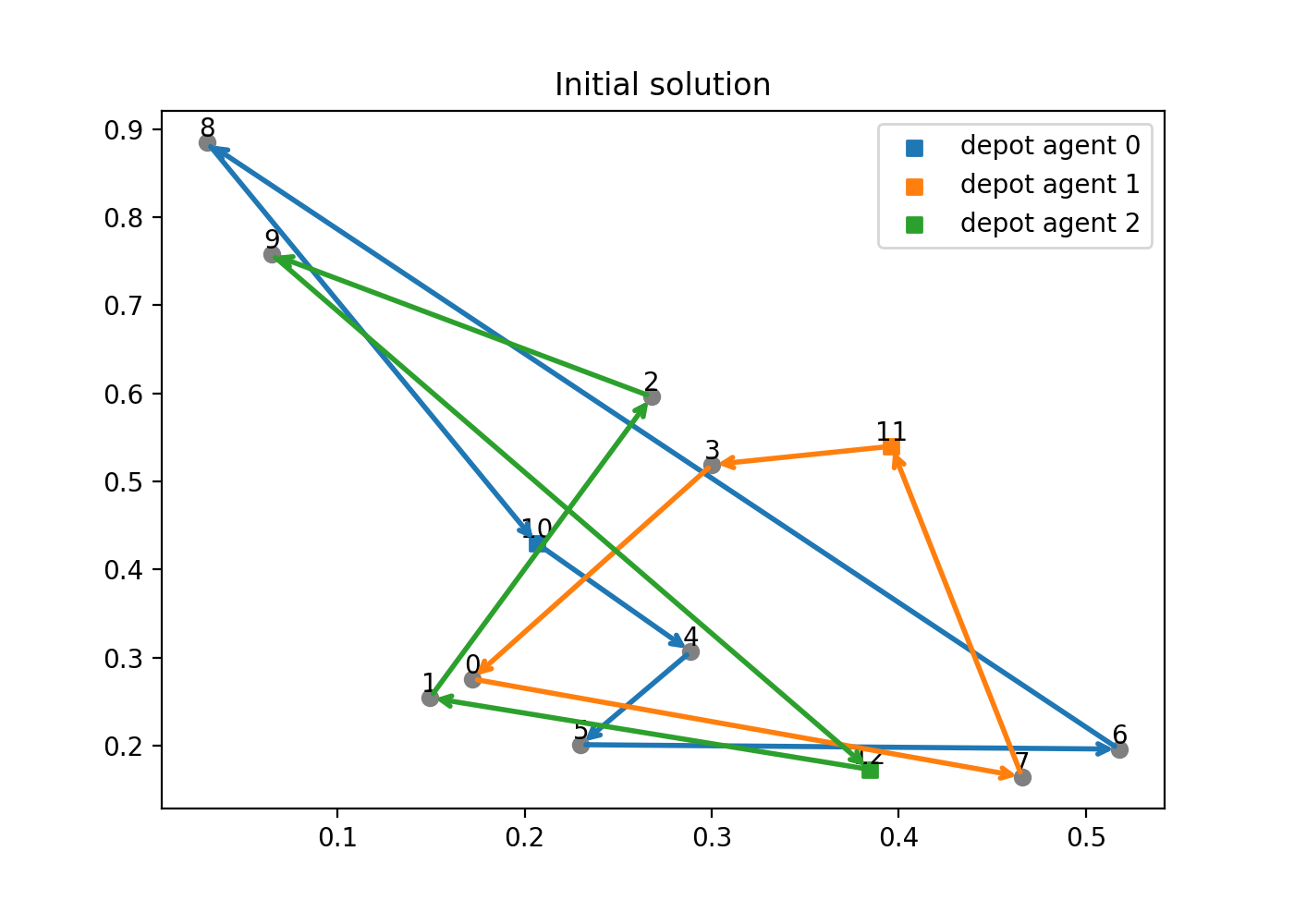}}%
        \raisebox{-\height}{\includegraphics[width=0.32\textwidth]{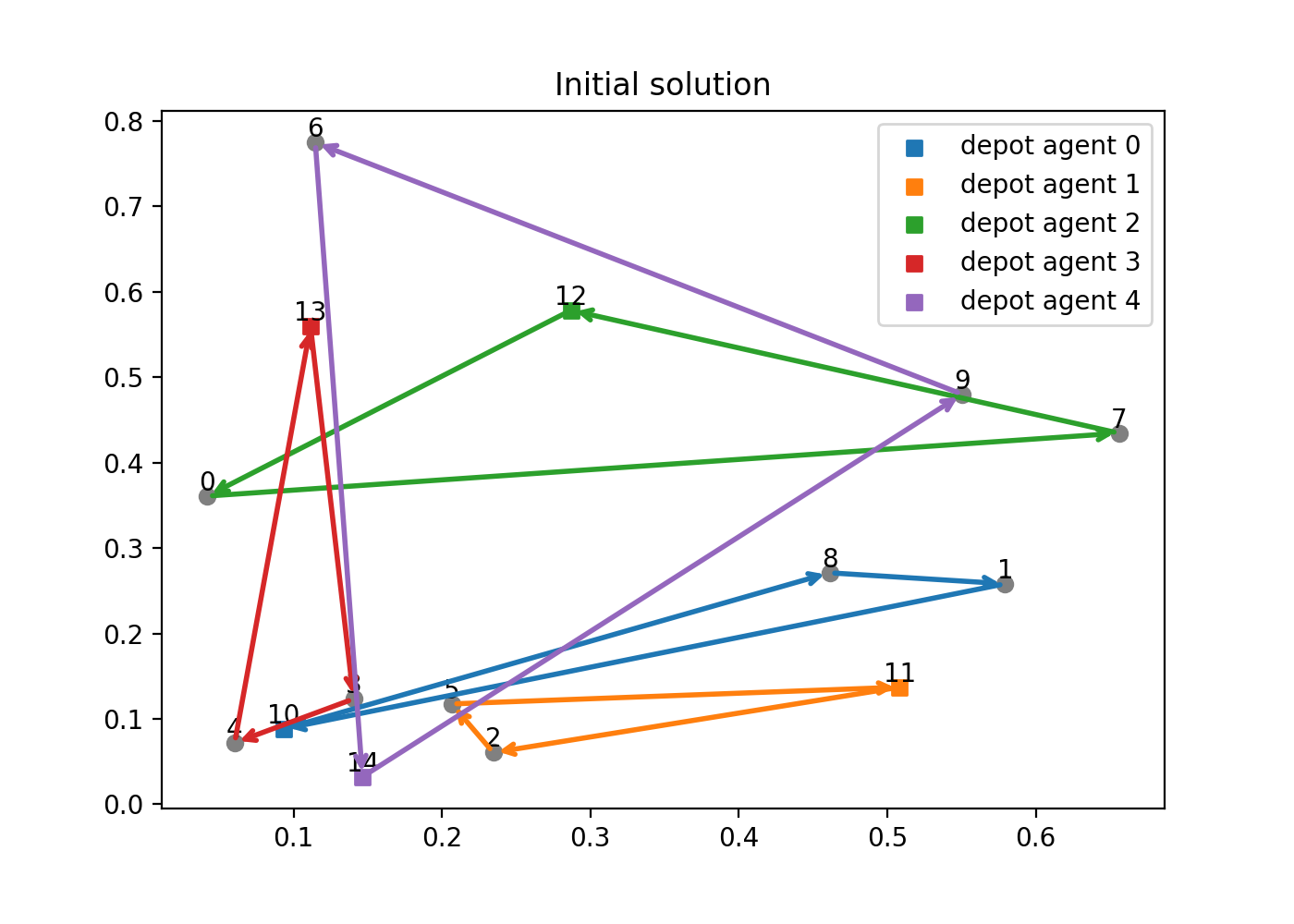}} %
    \end{subfigure}
    \caption{Exemplary sampled routing problems with corresponding sampled initial solutions for $10$ customers and a varying amount of agents. As nodes are randomly assigned to agents for the initial solution, chances are high that there is much room for improvement whereby agents will have to use the pool to exchange nodes.}
\end{figure}

\subsection{Hyperparameters}
\label{app_hyperparam}
Hyperparameter tuning was performed on the validation set with Ray Tune\footnote{https://docs.ray.io/en/latest/tune/}. Some hyperparameters could be set equally for all experiments, while others depend on the problem size or the number of agents.\newline
For a fixed problem size $\displaystyle k \in \{10,20\}$ we could use the same number of candidate actions $\displaystyle Z$ as well as the same amount of rewriting steps $\displaystyle T$.
Apart from one exception (see below) we could also use the same learning rates for $\displaystyle Q$ and for $\displaystyle \pi_u$ (controlled by $\alpha$ in \eqref{total_loss}) for a fixed problem size.\newline
For a fixed number of agents $n \in \{2,3,5\}$ we used the same value of $\displaystyle m$ in \eqref{reward} which manages the maximum number of allowed infeasible solutions in a row. We set $\displaystyle m = n + 1$ to allow each of the $\displaystyle n$ agents to decline the offered integration of a node from the pool once. The penalty is awarded as soon as one agent declines for a second time.\newline
See Table \ref{table_hyperparams} for a detailed overview about all experiment configurations.

\begin{table}[h]
  \centering
    \caption{Selected hyperparameter values for all experiments. The first block contains the hyperparameters which were set equally throughout all experiments.}
    \begin{tabular}{|c||c|c|c|c|c|c|}
    \hline
    \multirow{2}{*}{Hyperparameter} & \multicolumn{3}{c|}{10 nodes} & \multicolumn{3}{c|}{20 nodes}  \\
    \cline{2-7}
    & 2 agents  & 3 agents  & 5 agents & 2 agents  & 3 agents  & 5 agents \\ \hline
    discount factor $\gamma$ & 0.5 & 0.5 & 0.5 & 0.5 & 0.5 & 0.5  \\ 
    critic learning rate & 5e-04 & 5e-04 & 5e-04 & 5e-04 & 5e-04 & 5e-04 \\
    learning rate decay rate & 0.9 & 0.9 & 0.9 & 0.9 & 0.9 & 0.9 \\
    learning rate decay steps & 200 & 200 & 200 & 200 & 200 & 200 \\ 
    epsilon-greedy $\epsilon$ & 0.15 & 0.15 & 0.15 & 0.15 & 0.15 & 0.15 \\
    gradient clip & 0.05 & 0.05 & 0.05 & 0.05 & 0.05 & 0.05 \\ \hline
    rewriting steps $T$ & 30 & 30 & 30 & 40 & 40 & 40 \\
    num candidates actions $Z$ & 5 & 5 & 5 & 10 & 10 & 10 \\
    max num pool filled $m$ & 3 & 4 & 6 & 3 & 4 & 6 \\
    num trained epochs & 30 & 30 & 30 & 30 & 30 & 23 \\
    policy learning rate factor $\alpha$ & 1e-05 & 1e-05 & 1e-05 & 1e-06 & 1e-06 & 5e-06 \\ \hline
    \end{tabular}%
    \label{table_hyperparams}
\end{table}

\subsection{More details on the evaluation}
\label{app_eval_figures}
\begin{figure}[!htb]%
    \centering
    \subfloat[\centering 2 agents]{{\includegraphics[width=0.28\textwidth]{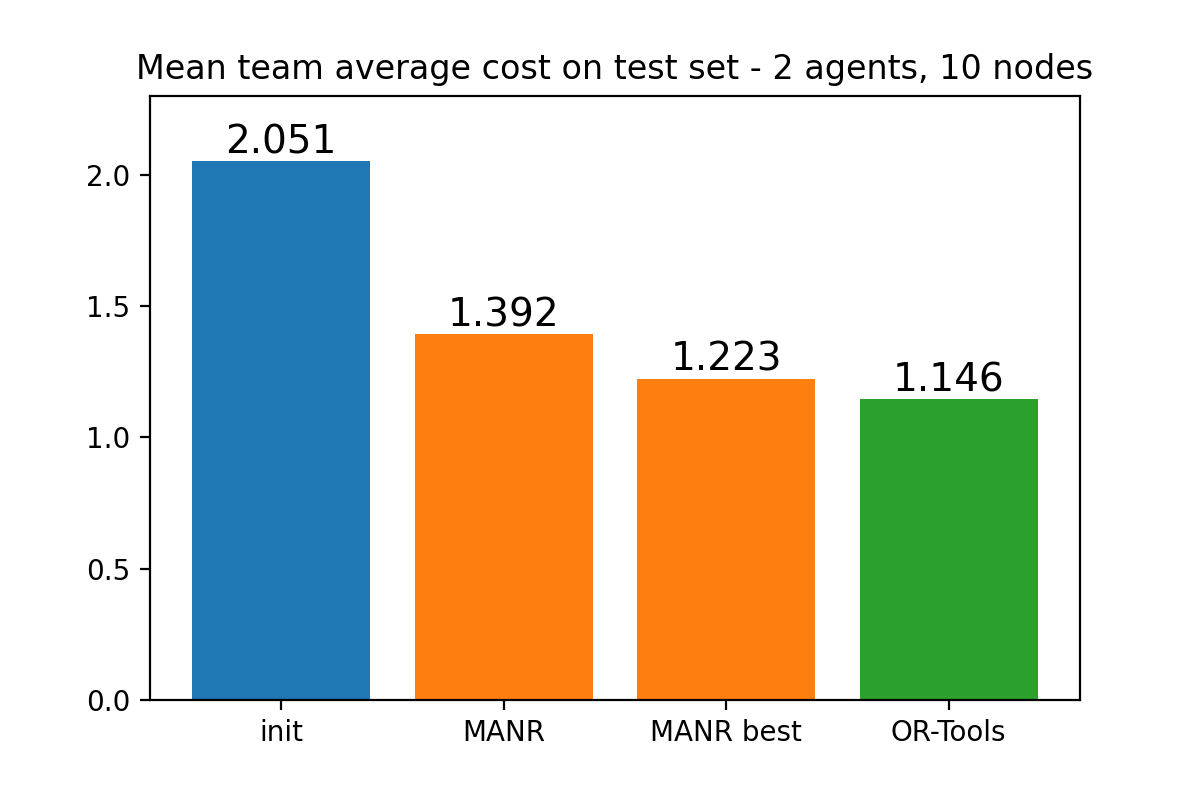} }}%
    \qquad
    \subfloat[\centering 3 agents]{{\includegraphics[width=0.28\textwidth]{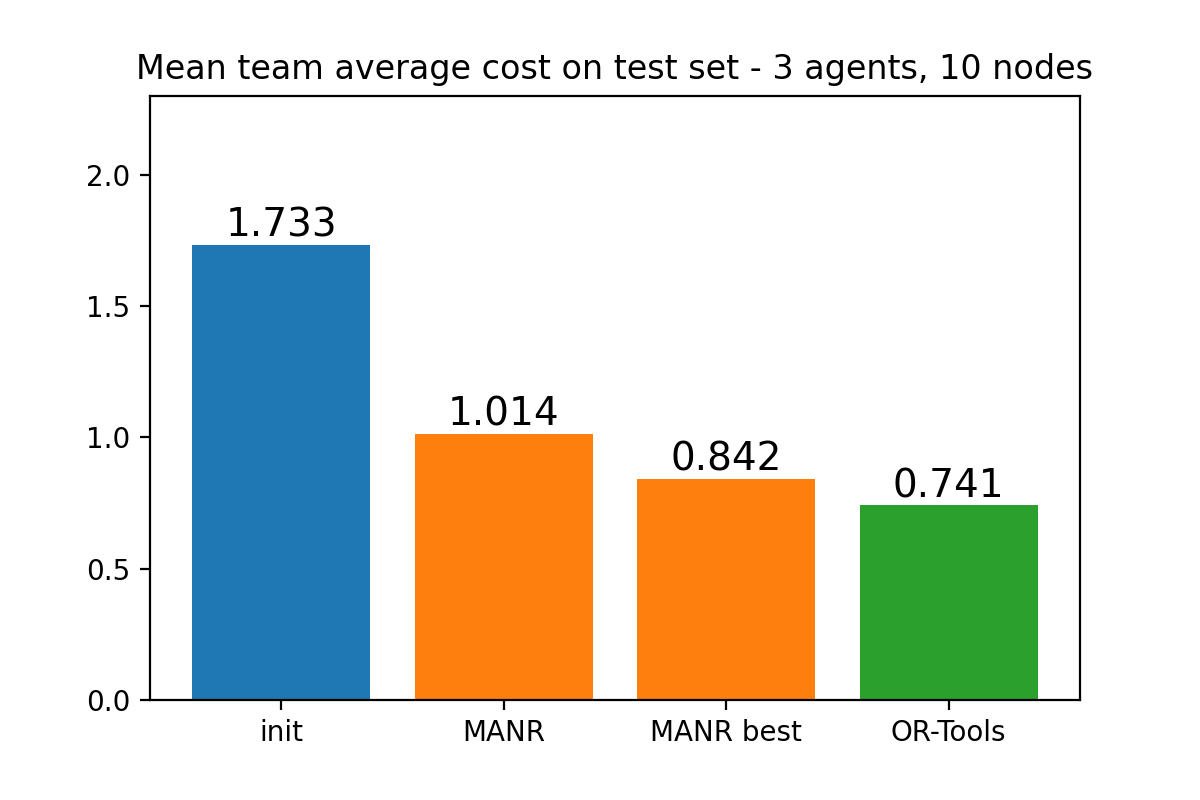} }}%
    \qquad
    \subfloat[\centering 5 agents]{{\includegraphics[width=0.28\textwidth]{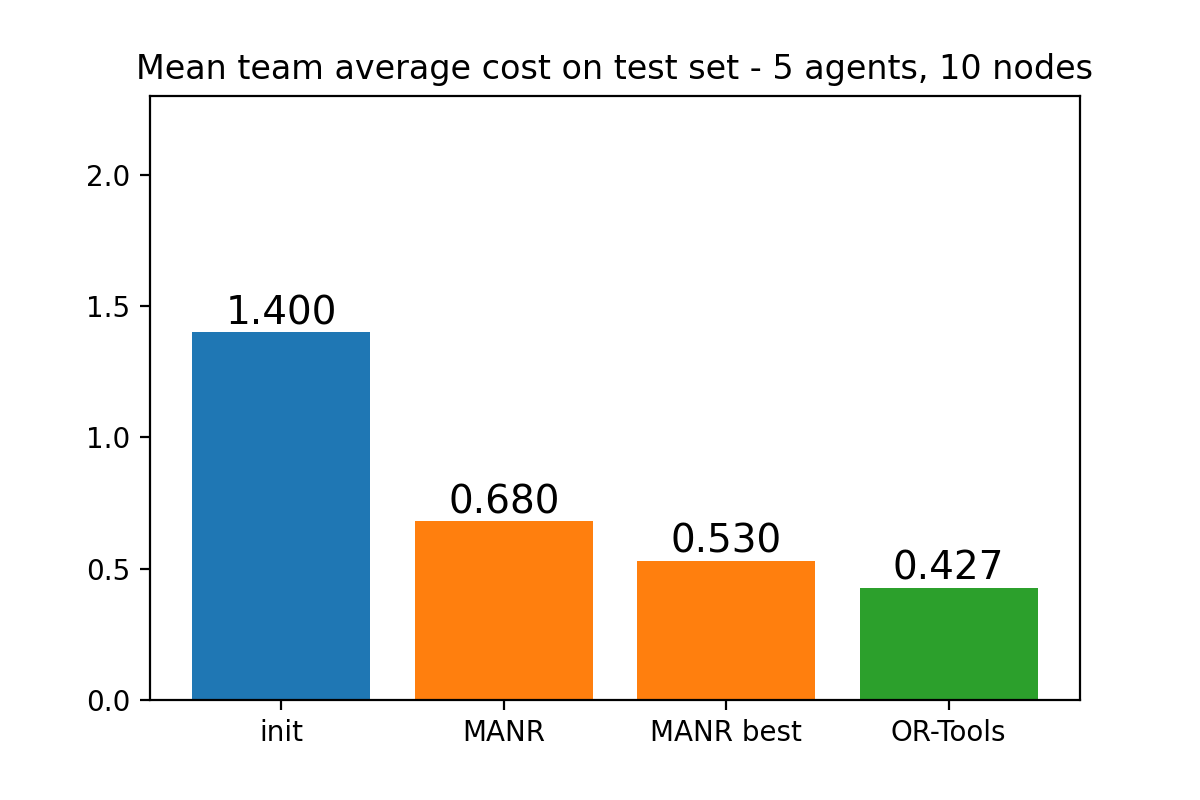} }}%
    \caption{Absolute performance values for 10 nodes and a varying amount of agents.}%
    \label{fig_eval_10}%
\end{figure}

\begin{figure}[!htb]%
    \centering
    \subfloat[\centering 2 agents]{{\includegraphics[width=0.28\textwidth]{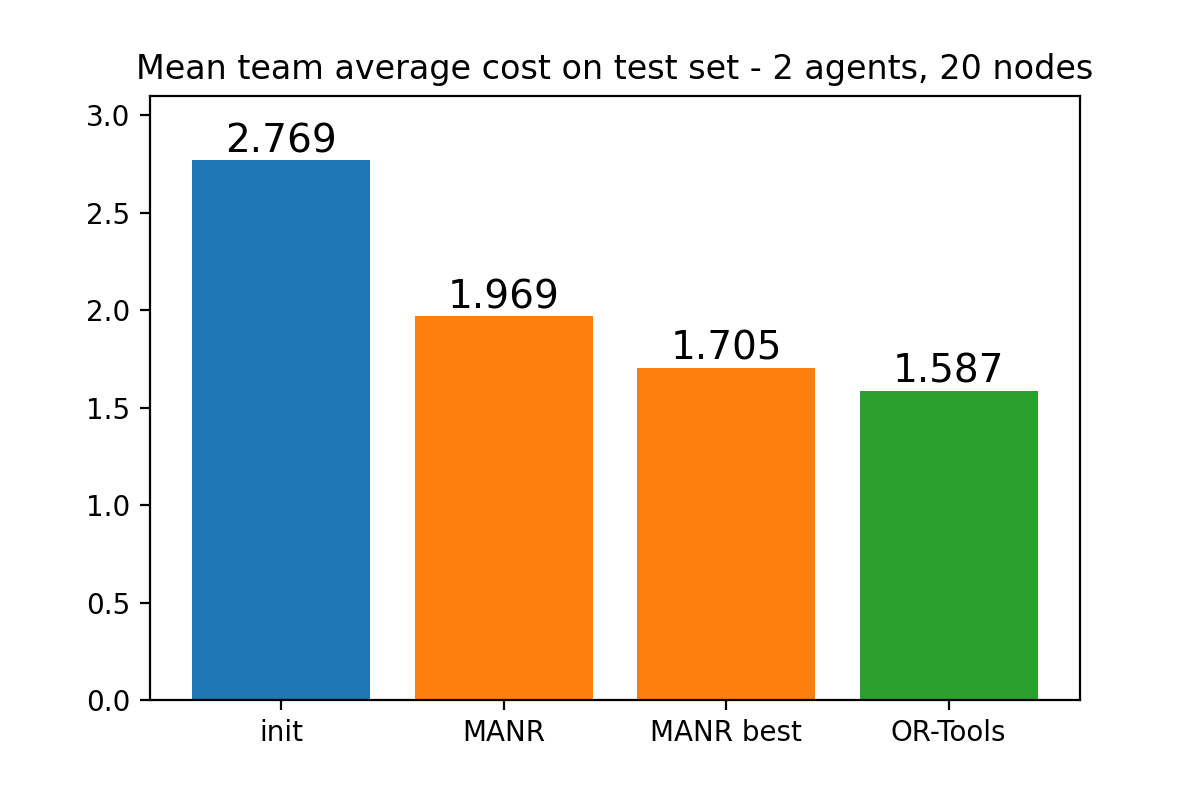} }}%
    \qquad
    \subfloat[\centering 3 agents]{{\includegraphics[width=0.28\textwidth]{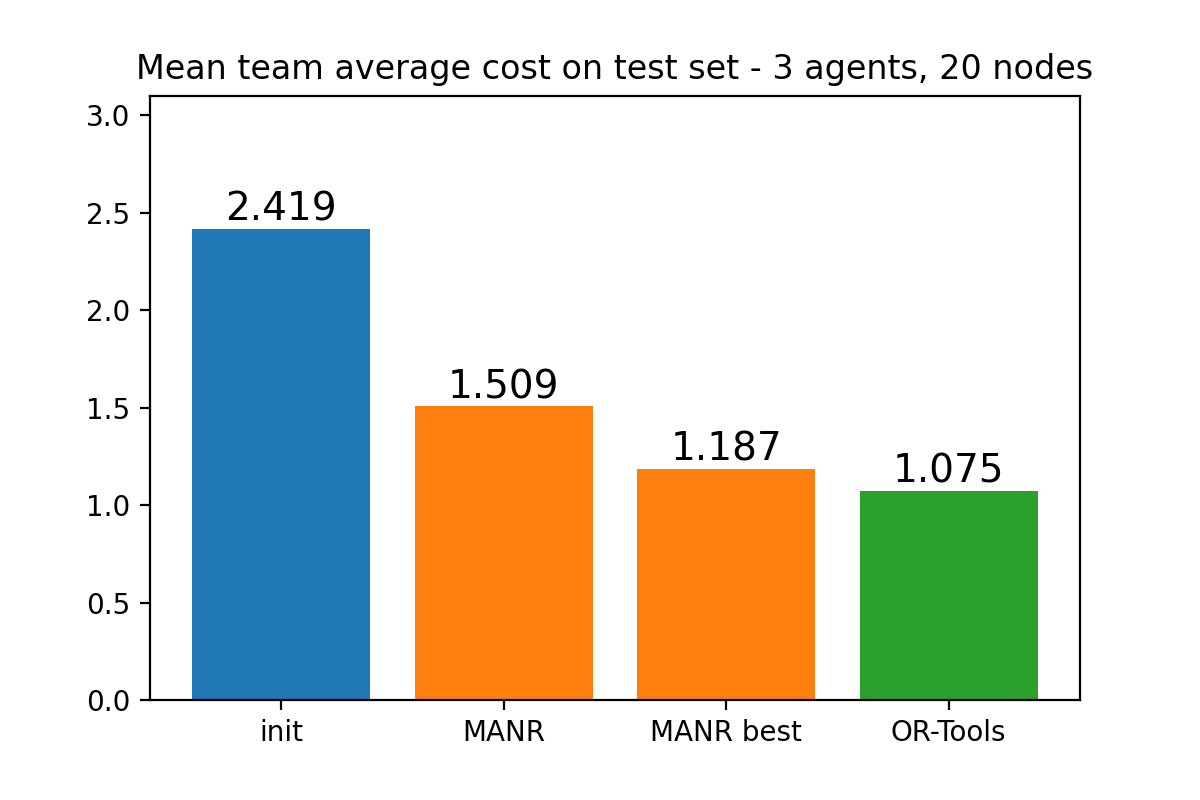} }}%
    \qquad
    \subfloat[\centering 5 agents]{{\includegraphics[width=0.28\textwidth]{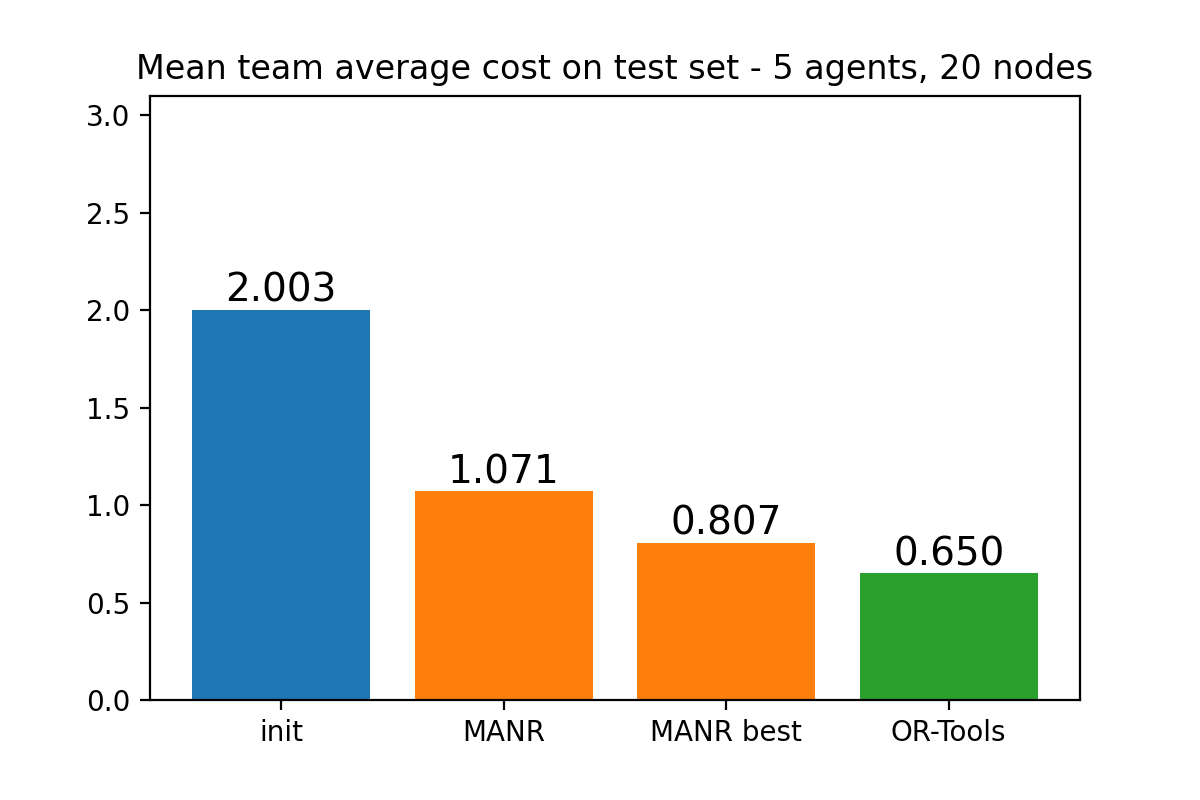} }}%
    \caption{Absolute performance values for 20 nodes and a varying amount of agents.}
    \label{fig_eval_20}%
\end{figure}

We report the performance gaps for the MANR, respectively MANR best, in terms of the percentage cost reduction relative to the intitial solution and relative to OR-Tools for the setup of $20$ customer nodes together with the corresponding run times in Table \ref{table_eval_20}.
Absolute performance values for the setups of $10$ and $20$ nodes are visualized in Figures \ref{fig_eval_10} and \ref{fig_eval_20} .

\begin{table}[!htb]
    \caption{Empirical results for $20$ customer nodes and a varying amount of agents.}
    \label{table_eval_20}
    \begin{subtable}{0.545\linewidth}
      \centering
        \caption{Performance gaps} 
        \begin{tabular}{|c||c|c|c|c|}
        \hline
        \multirow{2}{*}{Setup} & \multicolumn{2}{c|}{gap init} & \multicolumn{2}{c|}{gap OR-Tools}  \\
        \cline{2-5}
        & MANR  & MANR best  & MANR & MANR best \\ \hline
        2 agents & 29\% & 38\% & -24\% & -7\%  \\ 
        3 agents & 38\% & 51\% & -40\% & -10\%\\ 
        5 agents & 47\% & 60\% & -65\% & -24\%\\ \hline
        \end{tabular}%
    \end{subtable}
    \begin{subtable}{0.545\linewidth}
      \centering
        \caption{Average run time in seconds}
        \begin{tabular}{|c|| c| c|}
            \hline
          \multirow{2}{*}{Setup} & \multirow{2}{*}{MANR} & \multirow{2}{*}{OR-Tools} \\
          &&\\
             \hline
            2 agents & 0.45  &  0.03 \\ 
            3 agents  & 0.73  & 0.04 \\
            5 agents &  0.9  & 0.04  \\
            \hline
        \end{tabular}
    \end{subtable}%
\end{table}

\subsubsection{Exemplary Rewriting rollout}
\label{app_rollout}
We visualize a rewriting rollout of a routing problem with $10$ nodes and $3$ agents. We show the first $15$ global states. For the rest of the rewriting episode of $100$ steps, the MANR stayed in the last depicted solution.

\begin{figure}
     \centering
    \begin{subfigure}[t]{1\textwidth}
        \raisebox{-\height}{\includegraphics[width=0.32\textwidth]{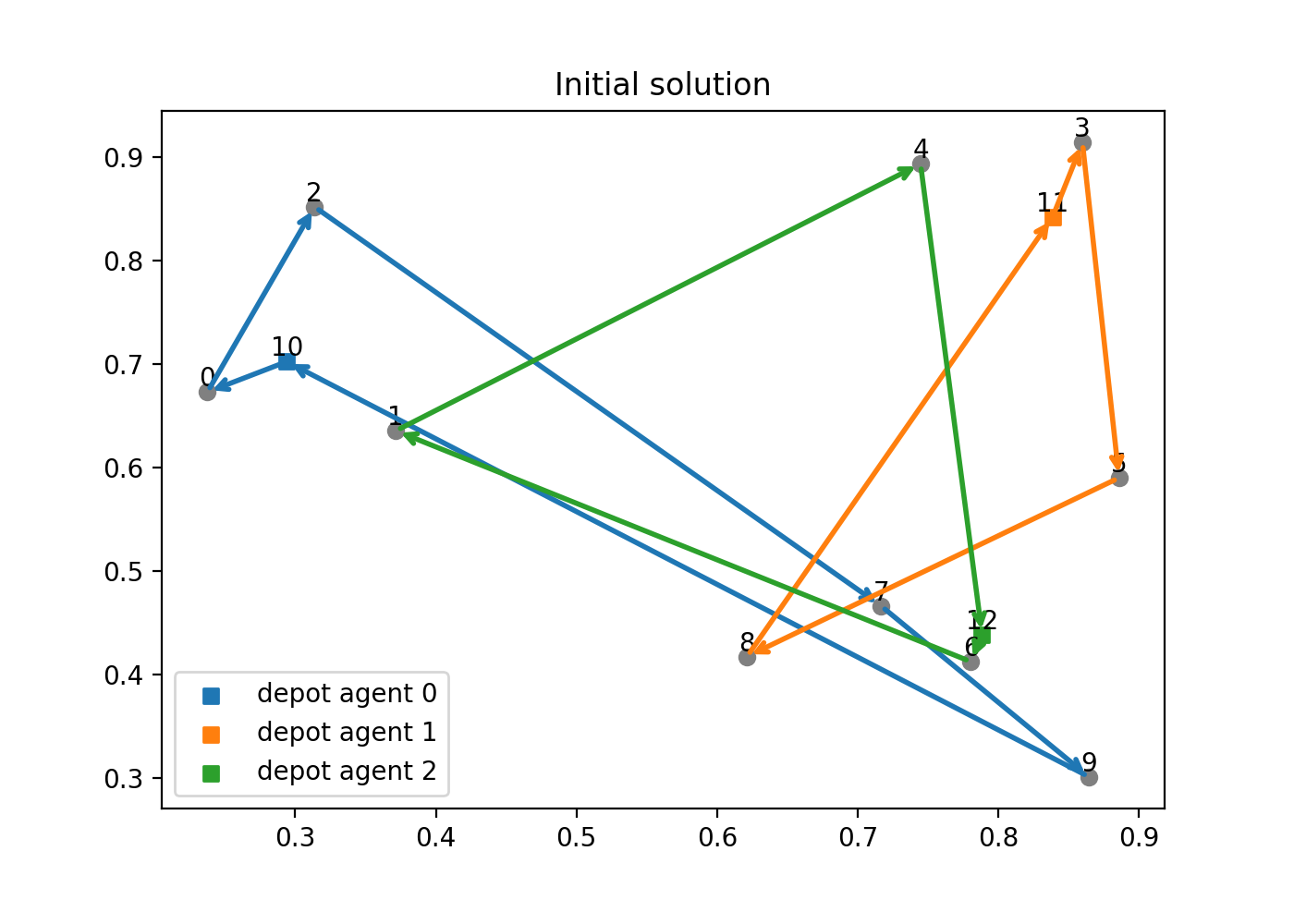}}%\hspace*{1em}
        \raisebox{-\height}{\includegraphics[width=0.32\textwidth]{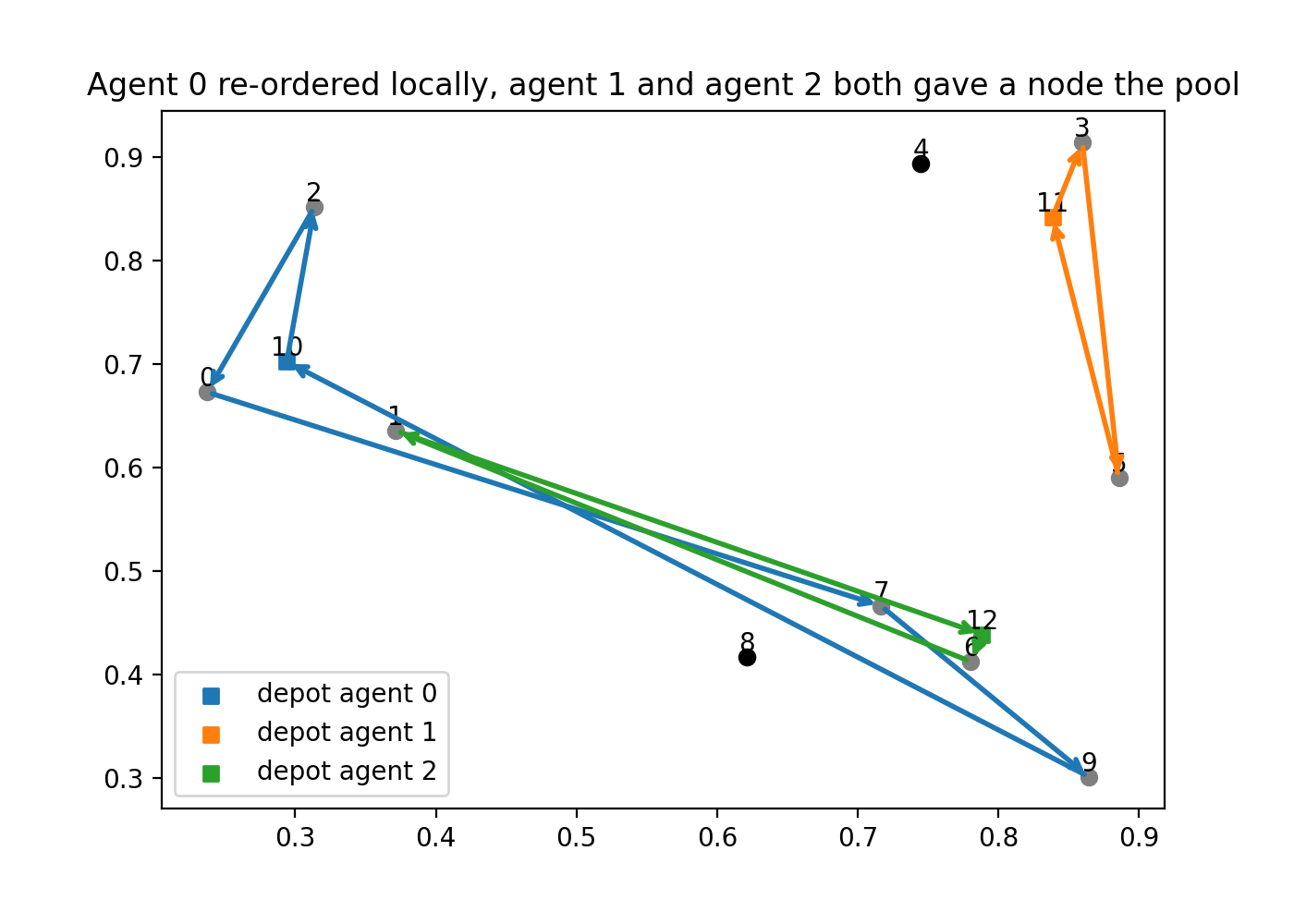}}%\hspace*{-3em}
        \raisebox{-\height}{\includegraphics[width=0.32\textwidth]{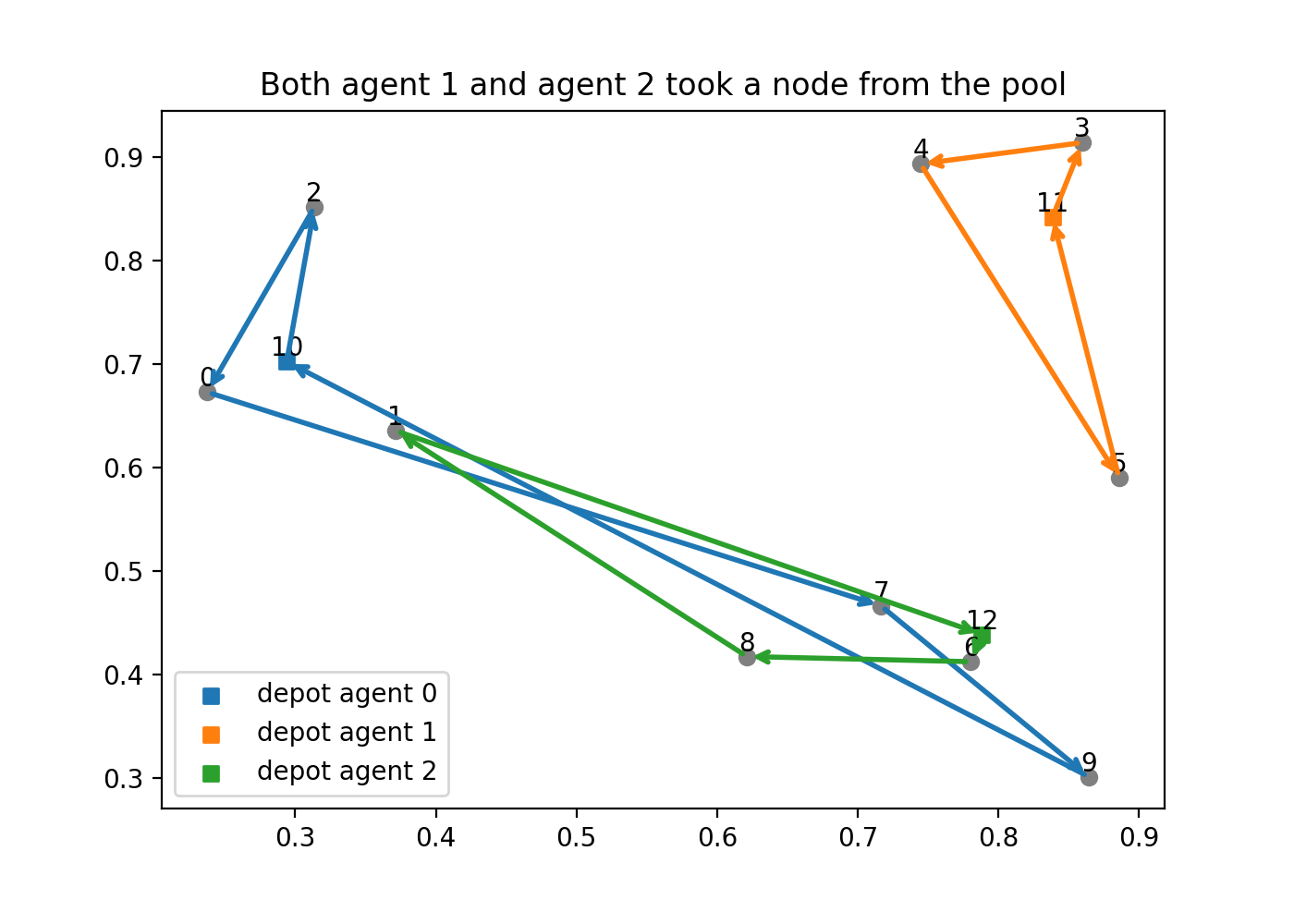}}%\hspace*{-3em}
    \end{subfigure}
    
    %%%%%%%%%%%%%%%%%%%%%%%%%%%%%%%%%%%% second row
    \begin{subfigure}[t]{1\textwidth}
        \raisebox{-\height}{\includegraphics[width=0.32\textwidth]{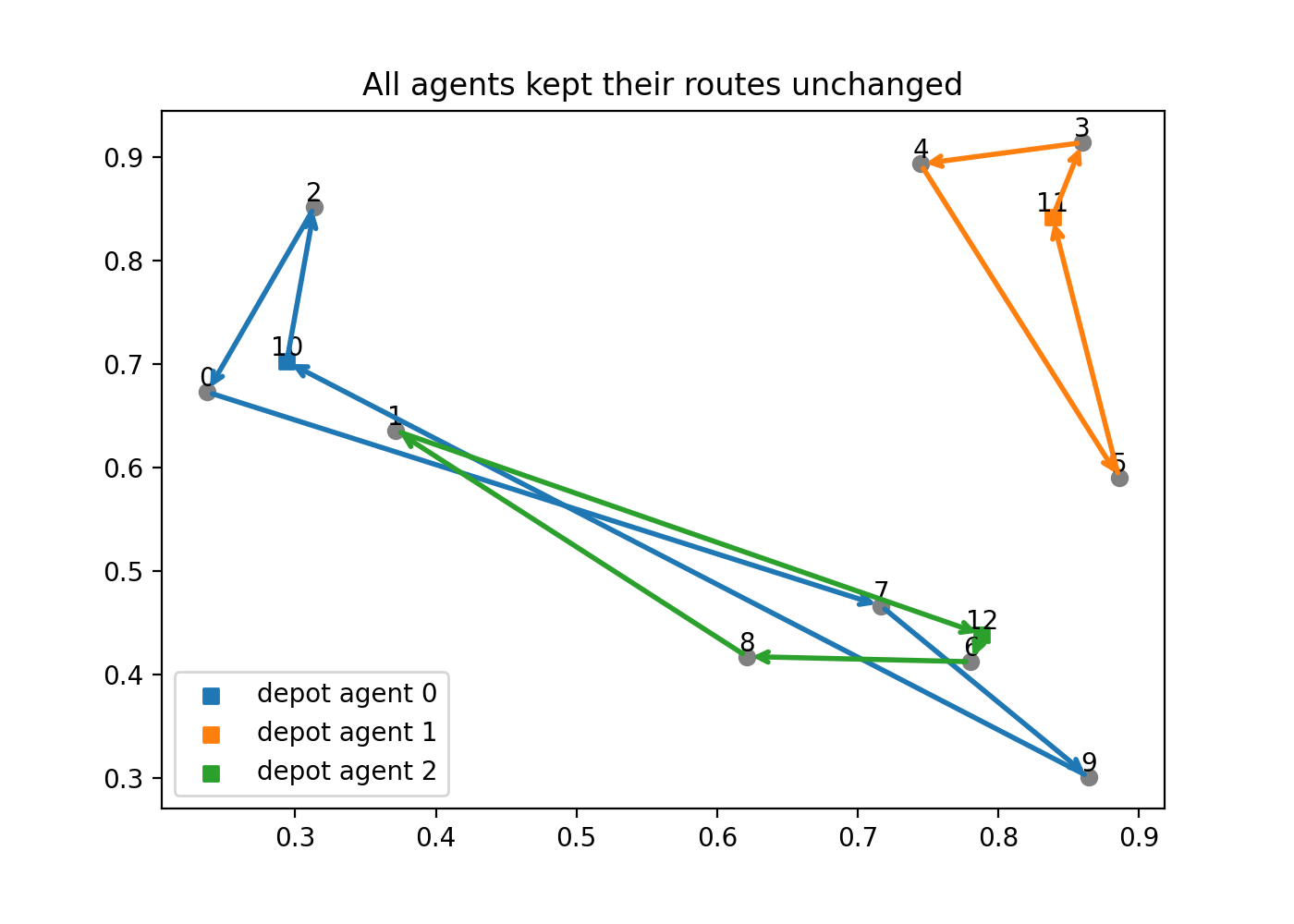}}%\hspace*{-3em}
        \raisebox{-\height}{\includegraphics[width=0.32\textwidth]{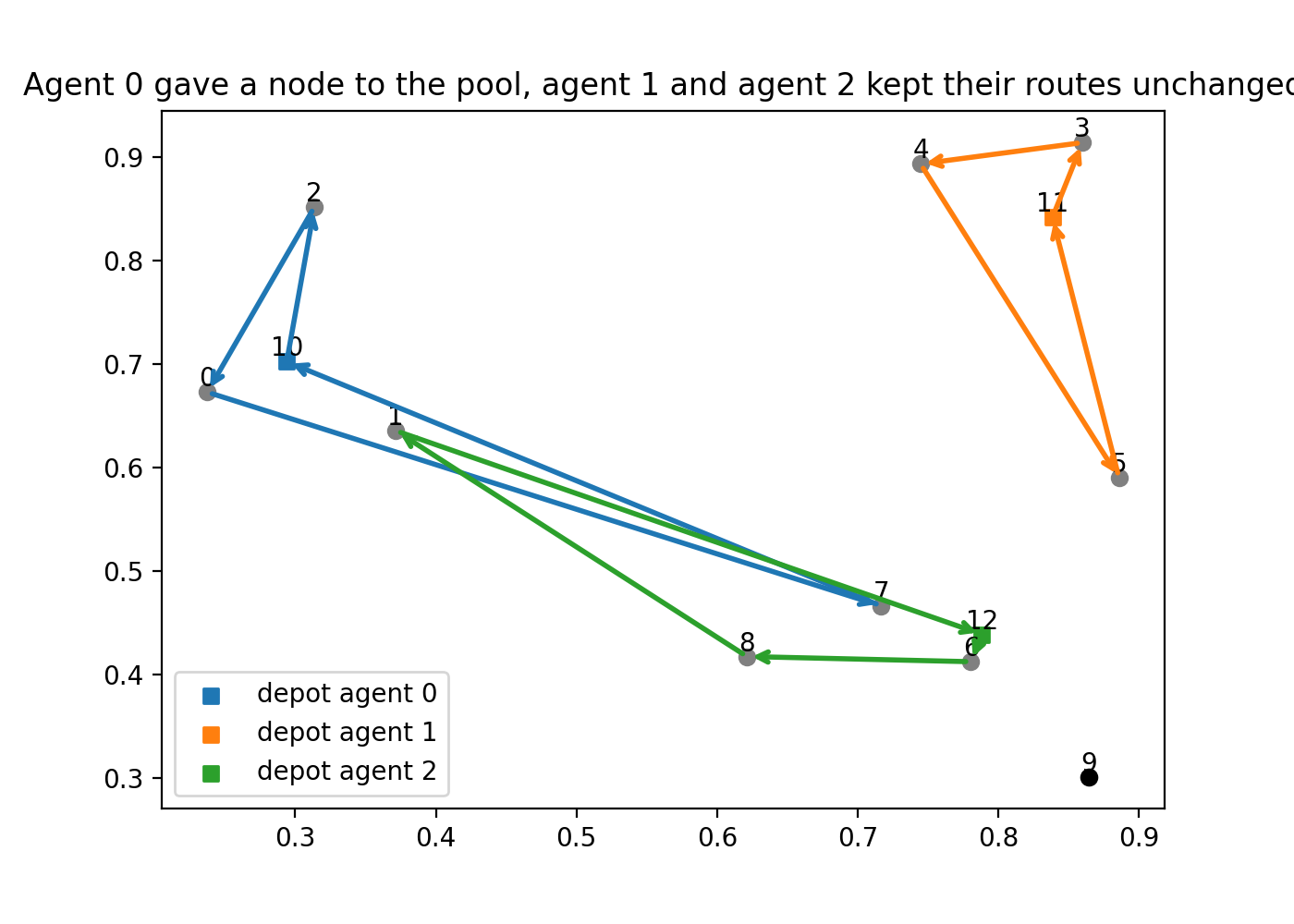}}%\hspace*{-3em}
        \raisebox{-\height}{\includegraphics[width=0.32\textwidth]{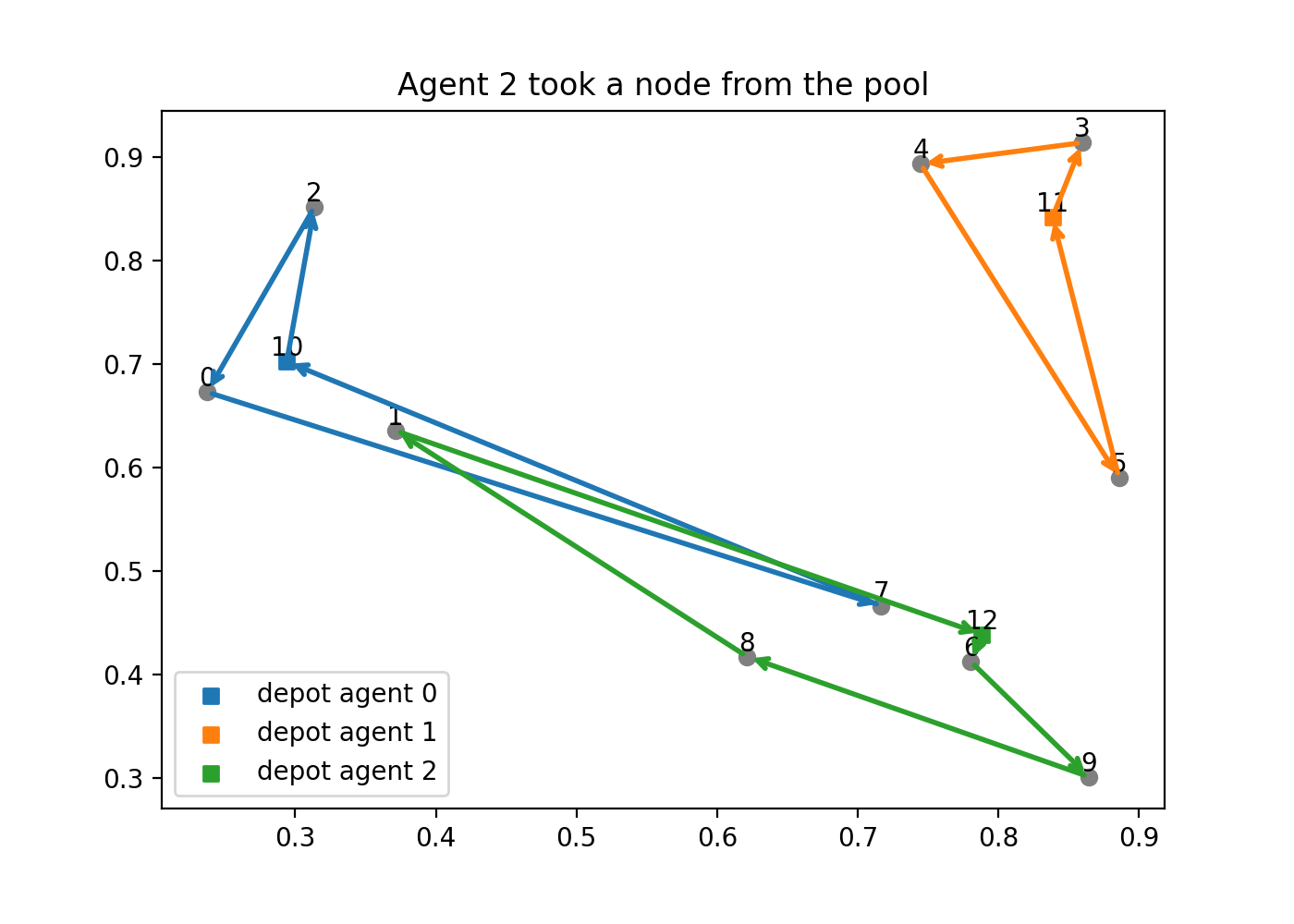}} %\hspace*{-3em}
    \end{subfigure}
    
    %%%%%%%%%%%%%%%%%%%%%%%%%%%%%%%%%%%% third row
    \begin{subfigure}[t]{1\textwidth}
        \raisebox{-\height}{\includegraphics[width=0.32\textwidth]{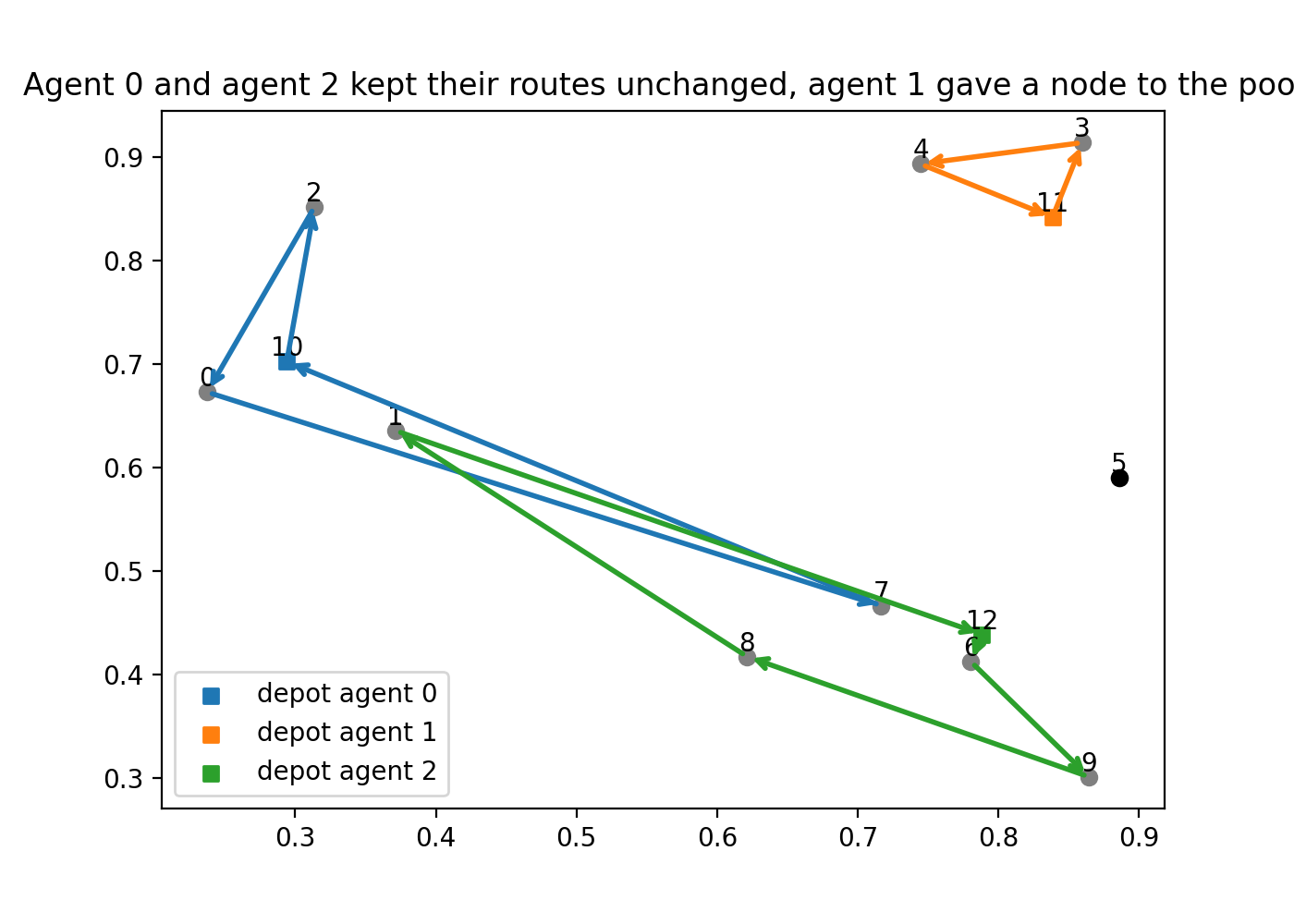}}%\hspace*{-3em}
        \raisebox{-\height}{\includegraphics[width=0.32\textwidth]{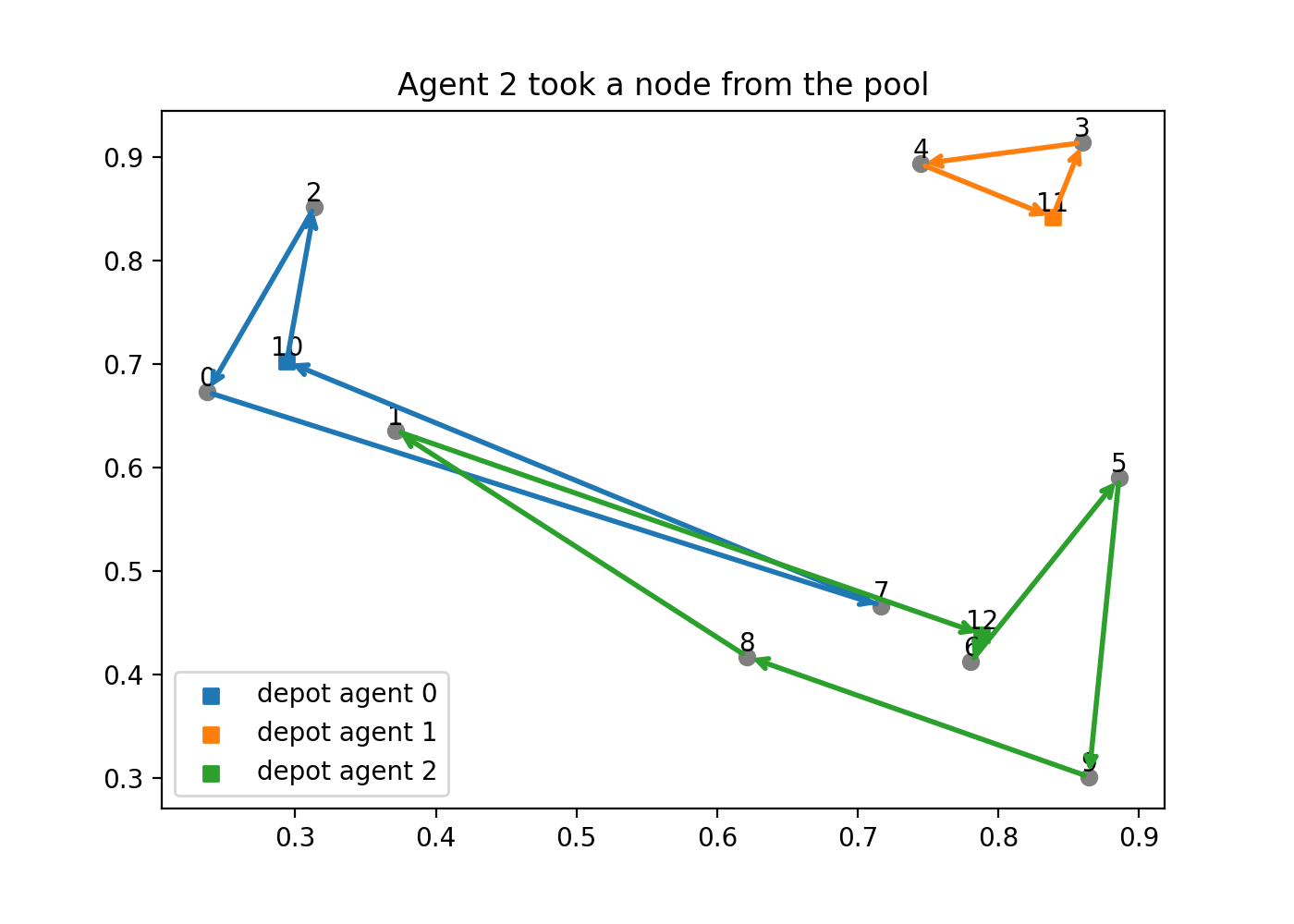}}%\hspace*{-3em}
        \raisebox{-\height}{\includegraphics[width=0.32\textwidth]{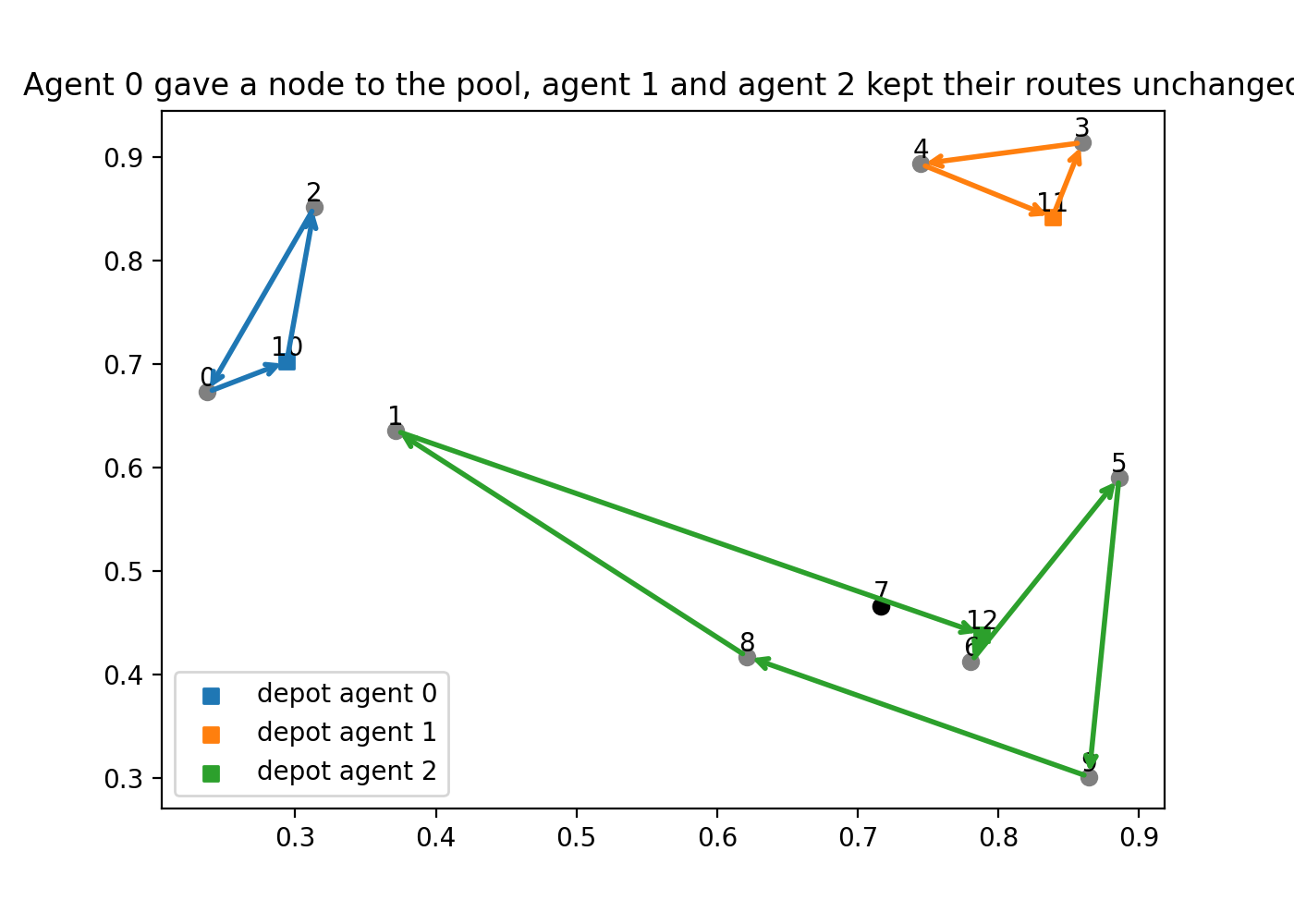}} %\hspace*{-3em}
    \end{subfigure}
    
    %%%%%%%%%%%%%%%%%%%%%%%%%%%%%%%%%%%% fourth row
    \begin{subfigure}[t]{1\textwidth}
        \raisebox{-\height}{\includegraphics[width=0.32\textwidth]{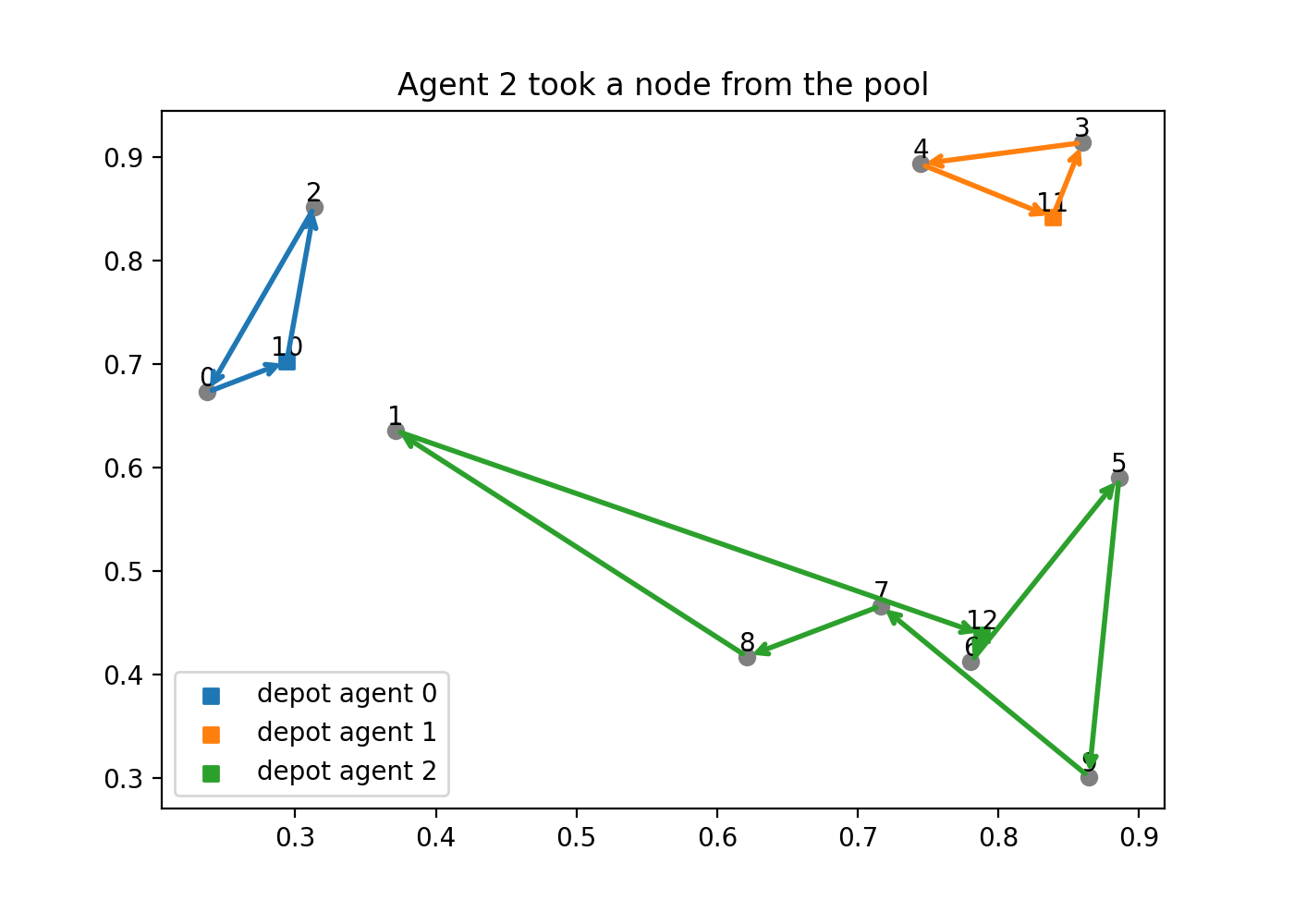}}%\hspace*{-3em}
        \raisebox{-\height}{\includegraphics[width=0.32\textwidth]{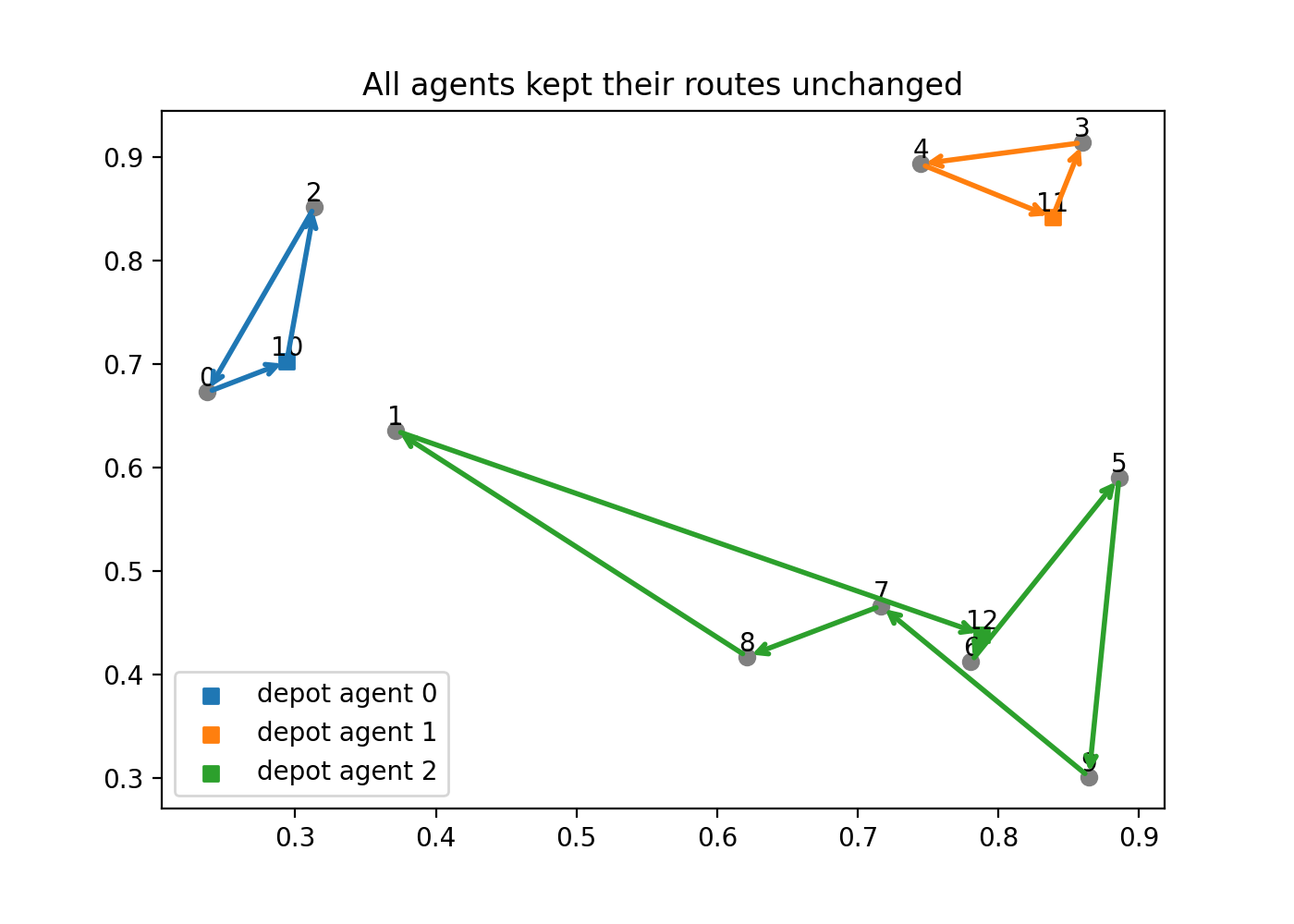}}%\hspace*{-3em}
        \raisebox{-\height}{\includegraphics[width=0.32\textwidth]{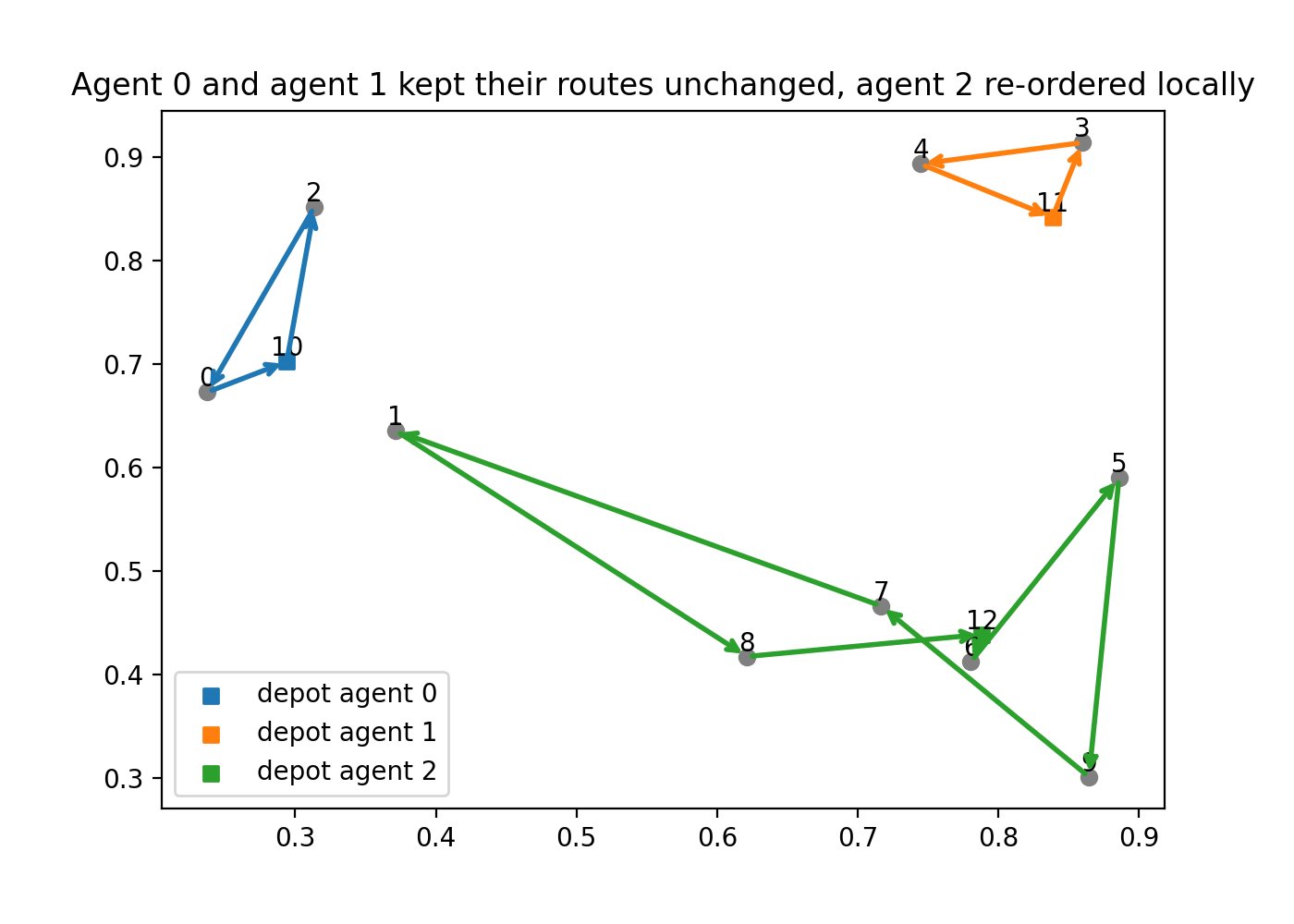}} %\hspace*{-3em}
    \end{subfigure}
    
    %%%%%%%%%%%%%%%%%%%%%%%%%%%%%%%%%%%% fifth row
    \begin{subfigure}[t]{1\textwidth}
        \raisebox{-\height}{\includegraphics[width=0.32\textwidth]{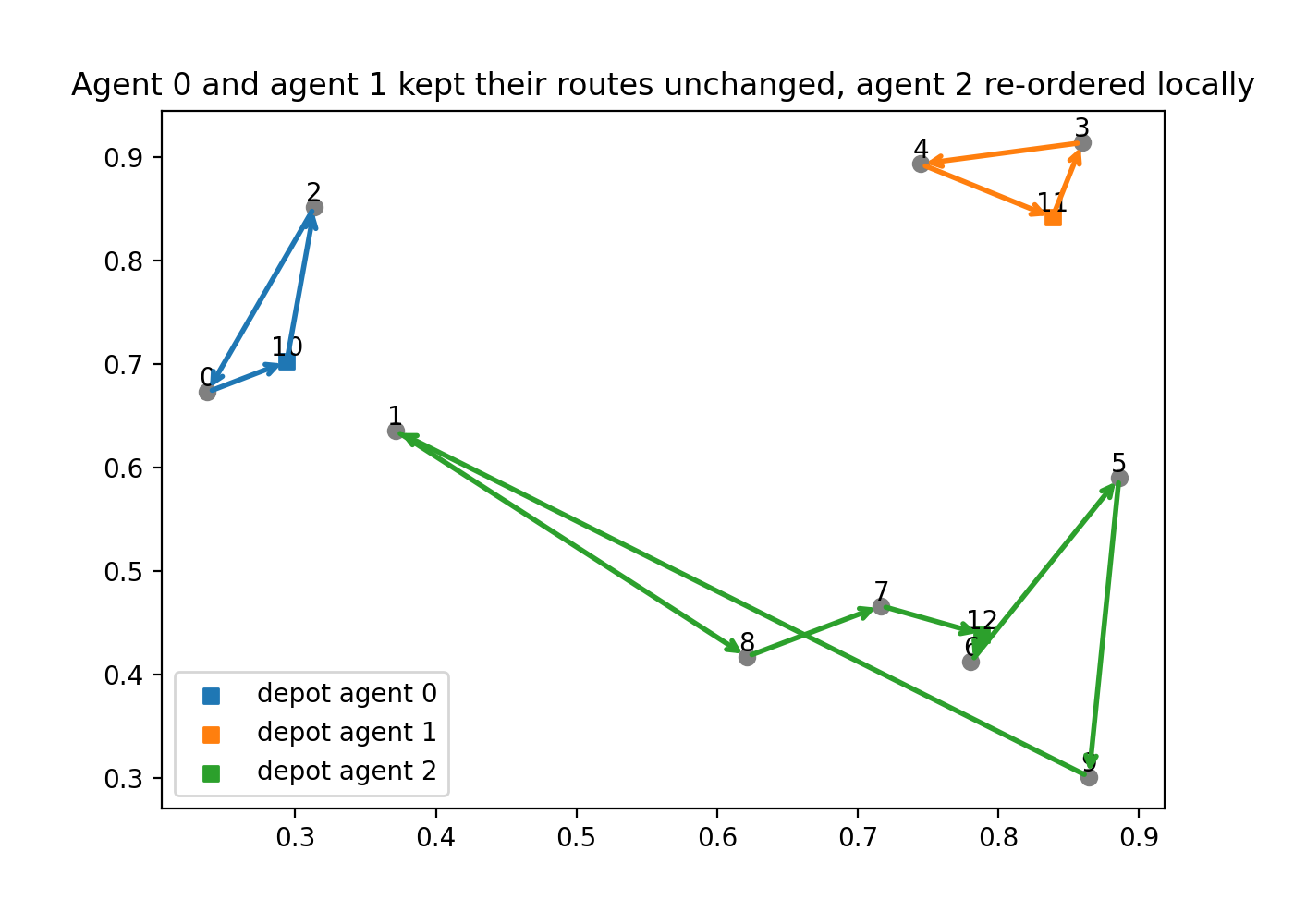}}%\hspace*{-3em}
        \raisebox{-\height}{\includegraphics[width=0.32\textwidth]{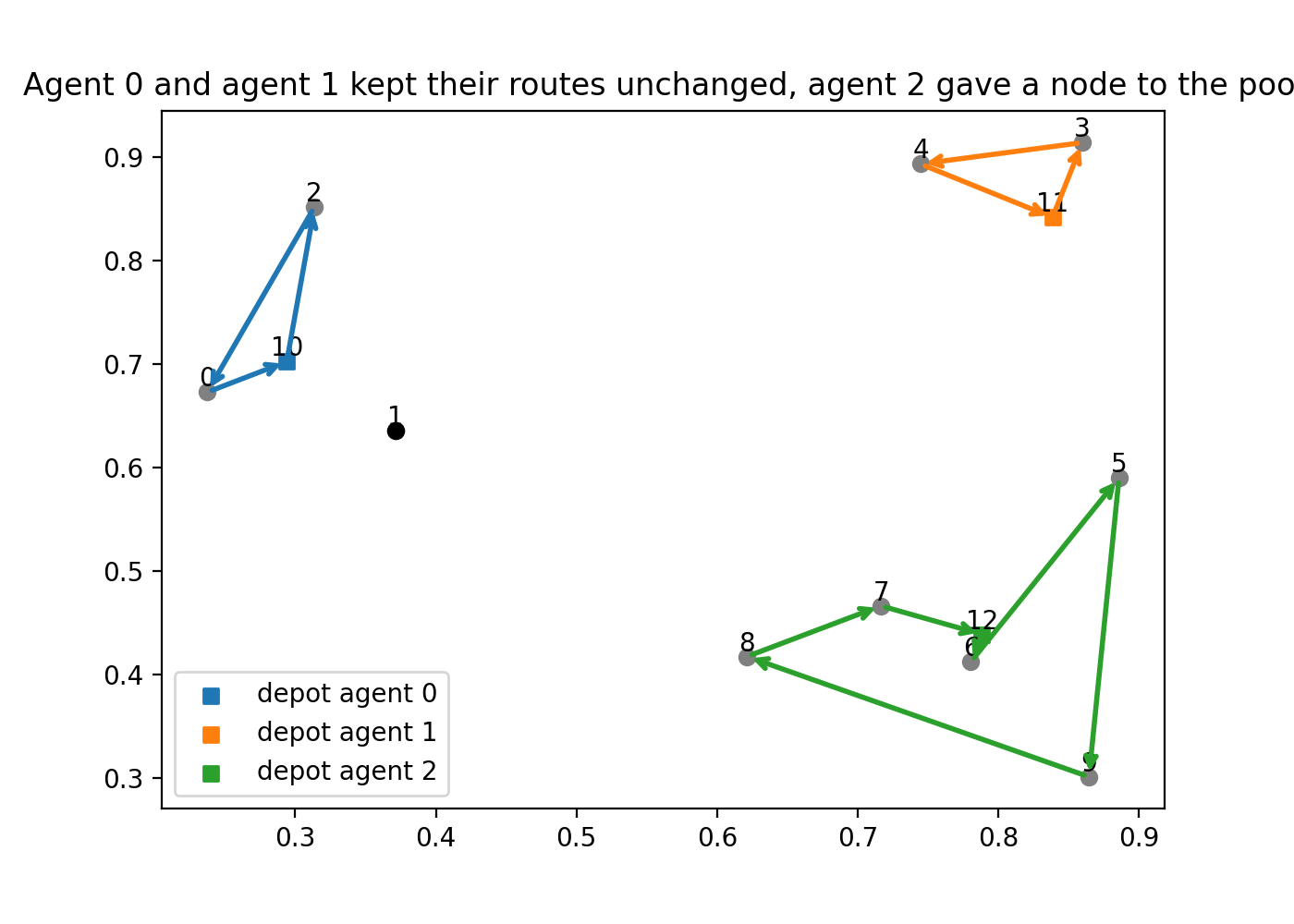}}%\hspace*{-3em}
        \raisebox{-\height}{\includegraphics[width=0.32\textwidth]{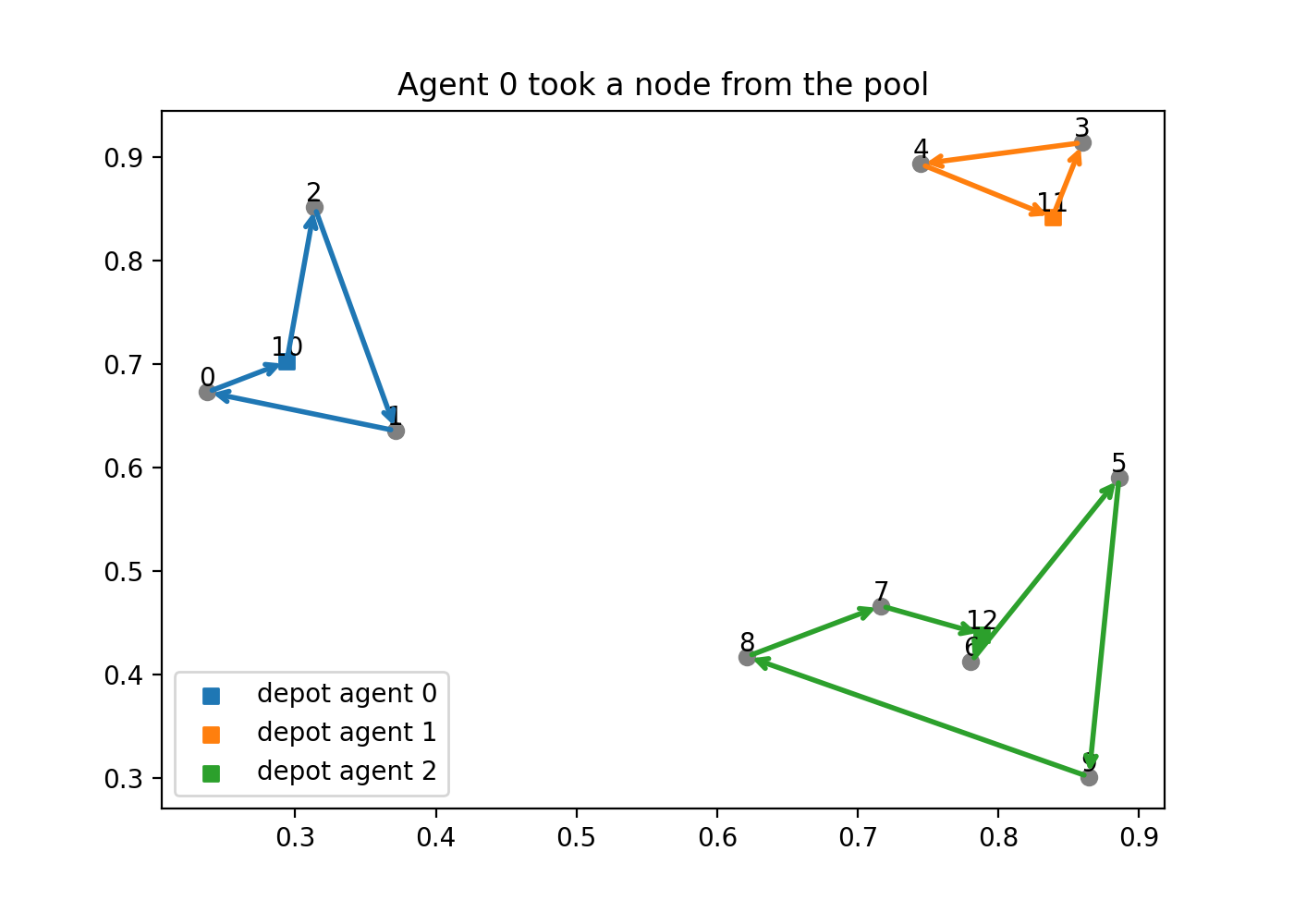}} %\hspace*{-3em}
    \end{subfigure}
    \caption{Exemplary excerpt from a rewriting episode. Note that an action of "doing nothing" is often caused by the randomly selected region node to be moved.}
    \label{fig_rollout}
\end{figure}

\end{document}